\documentclass[Afour,sageh,times]{sagej}

\def\volumeyear{2023}

\usepackage{amsmath}
\usepackage{amsfonts}
\usepackage{amssymb}
\usepackage{bm}
\usepackage{mathtools}
\usepackage{amsthm}
\usepackage{caption}
\usepackage{subcaption}
\usepackage{algorithm}
\usepackage{algpseudocode}
\usepackage[colorlinks,bookmarksopen,bookmarksnumbered,citecolor=red,urlcolor=red]{hyperref}
\usepackage{color}
\usepackage{tikz}
\usetikzlibrary{decorations.pathreplacing, positioning, arrows.meta}
\usepackage{tabularx}
\usepackage{arydshln}
\usepackage{threeparttable} 
\usepackage{diagbox}

\newcommand{\q}{\bm{q}}
\renewcommand{\v}{\bm{v}}
\renewcommand{\a}{\bm{a}}

\renewcommand{\u}{\bm{u}}
\newcommand{\x}{\bm{x}}

\newcommand{\M}{\bm{M}}
\newcommand{\C}{\bm{k}}
\newcommand{\B}{\bm{B}}
\newcommand{\btau}{\bm{\tau}}
\newcommand{\J}{\bm{J}}
\newcommand{\f}{\bm{f}}

\newcommand{\N}{\bm{N}}

\newcommand{\Q}{\bm{Q}}
\newcommand{\R}{\bm{R}}

\renewcommand{\H}{\bm{H}}
\newcommand{\p}{\bm{p}}
\newcommand{\g}{\bm{g}}
\newcommand{\h}{\bm{h}}

\newcommand{\A}{\bm{A}}
\newcommand{\D}{\bm{D}}
\newcommand{\blambda}{\bm{\lambda}}

\newcommand{\K}{\bm{K}}

\theoremstyle{plain}
\newtheorem{remark}{Remark}

\newcommand\clearrow{\global\let\rowmac\relax}
\clearrow

\newcommand{\hl}[1]{#1}    

\newcommand{\msf}[1]{$\mathsf{#1}$}

\begin{document}

\runninghead{Kurtz et al.}

\title{Inverse Dynamics Trajectory Optimization for Contact-Implicit Model Predictive Control}

\author{Vince Kurtz\affilnum{1,3}, Alejandro Castro\affilnum{2}, Aykut \"{O}zg\"{u}n \"{O}nol\affilnum{2}, and Hai Lin\affilnum{3}}

\affiliation{\affilnum{1}Department of Civil and Mechanical Engineering, California Institute of Technology, USA \\
    \affilnum{2}Toyota Research Institute, USA \\
    \affilnum{3}Department of Electrical Engineering, University of Notre Dame, USA
}

\corrauth{Vince Kurtz, California Institute of Technology, 1200 E California Blvd, Pasadena, CA 91125, USA}
\email{vkurtz@caltech.edu}


\begin{abstract}
Robots must make and break contact with the environment to perform useful tasks,
but planning and control through contact remains a formidable challenge. In this
work, we achieve real-time contact-implicit model predictive control with a
surprisingly simple method: inverse dynamics trajectory optimization. While
trajectory optimization with inverse dynamics is not new, we introduce a series
of incremental innovations that collectively enable fast model predictive
control on a variety of challenging manipulation and locomotion tasks. We
implement these innovations in an open-source solver and present simulation
examples to support the effectiveness of the proposed approach. Additionally, we
demonstrate contact-implicit model predictive control on hardware at over 100~Hz
for a 20-degree-of-freedom bi-manual manipulation task. Video and code are
available at \url{https://idto.github.io}.
\end{abstract}

\maketitle

\section{Introduction}

Contact is critical for legged locomotion and dexterous manipulation, but most
optimization-based controllers assume a fixed contact sequence
\citep{wensing2022optimization}. Contact-Implicit Trajectory Optimization (CITO)
aims to relax this assumption by solving jointly for the contact sequence and a
continuous motion. A sufficiently performant CITO solver would enable
Contact-Implicit Model Predictive Control (CI-MPC), allowing robots to determine
contact modes on the fly and perform more complex tasks.

Despite growing interest in CITO, fast and reliable CI-MPC remains elusive. Most
existing methods take either a direct approach \citep{posa2013direct,
manchester2019contact, winkler2018gait, patel2019contact, moura2022non,
wang2023contact, aydinoglu2023consensus,cleac2023fast}, in which decision
variables represent state and control at each time step, or a shooting approach
\citep{tassa2012synthesis, neunert2017trajectory, carius2018trajectory,
chatzinikolaidis2021trajectory,kim2022contact, howell2022trajectory,
kong2022hybrid}, where control inputs are the only decision variables. Both work
well for smooth dynamical systems, but struggle to handle contact. The large
number of additional constraints used to model contact, rank deficiency, and ill
conditioning have a large impact on numerics. Poor numerics in turn degrade
robustness, convergence, and performance. Other issues include the difficulty of
initializing and/or warm-starting variables, locally optimal solutions that do
not obey physics, and high sensitivity to problem parameters. Much active
research focuses on solving these problems. 

\begin{figure}
    \centering
    \begin{subfigure}{0.49\linewidth}
        \centering
        \includegraphics[width=0.95\linewidth]{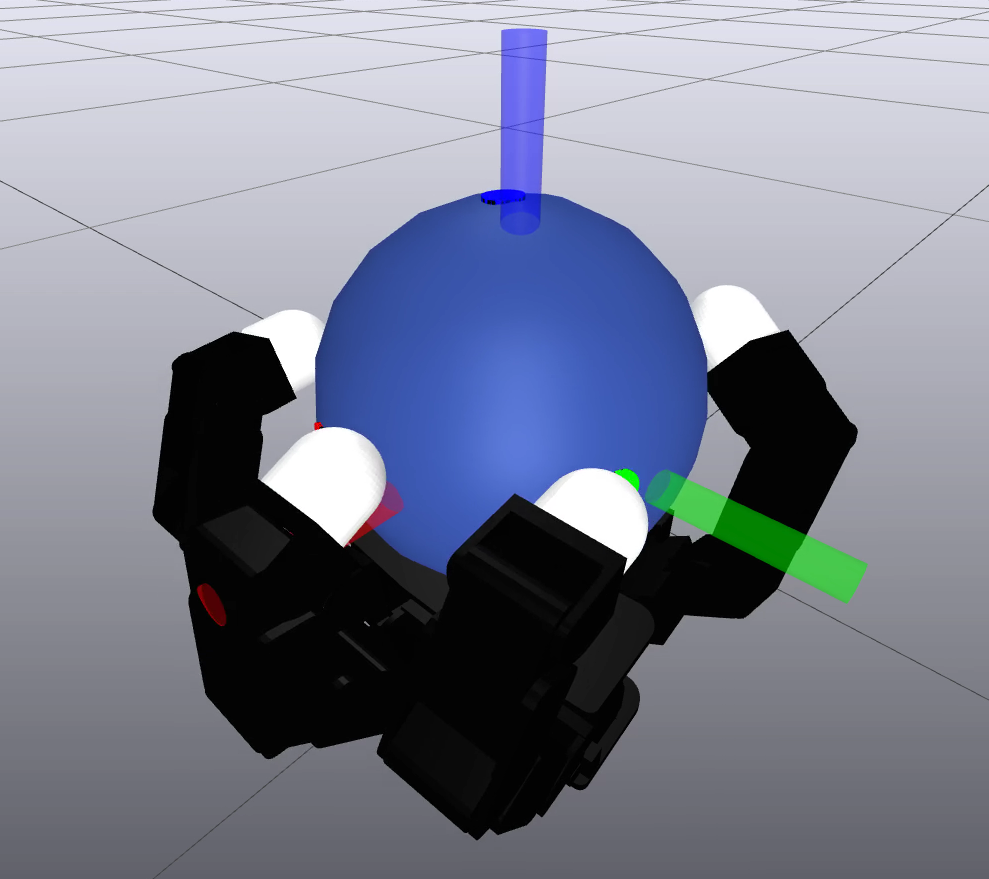}
        \caption{Dexterous Hand: 10~Hz}
        \label{fig:cover:allegro}
    \end{subfigure}
    \begin{subfigure}{0.49\linewidth}
        \centering
        \includegraphics[width=0.95\linewidth]{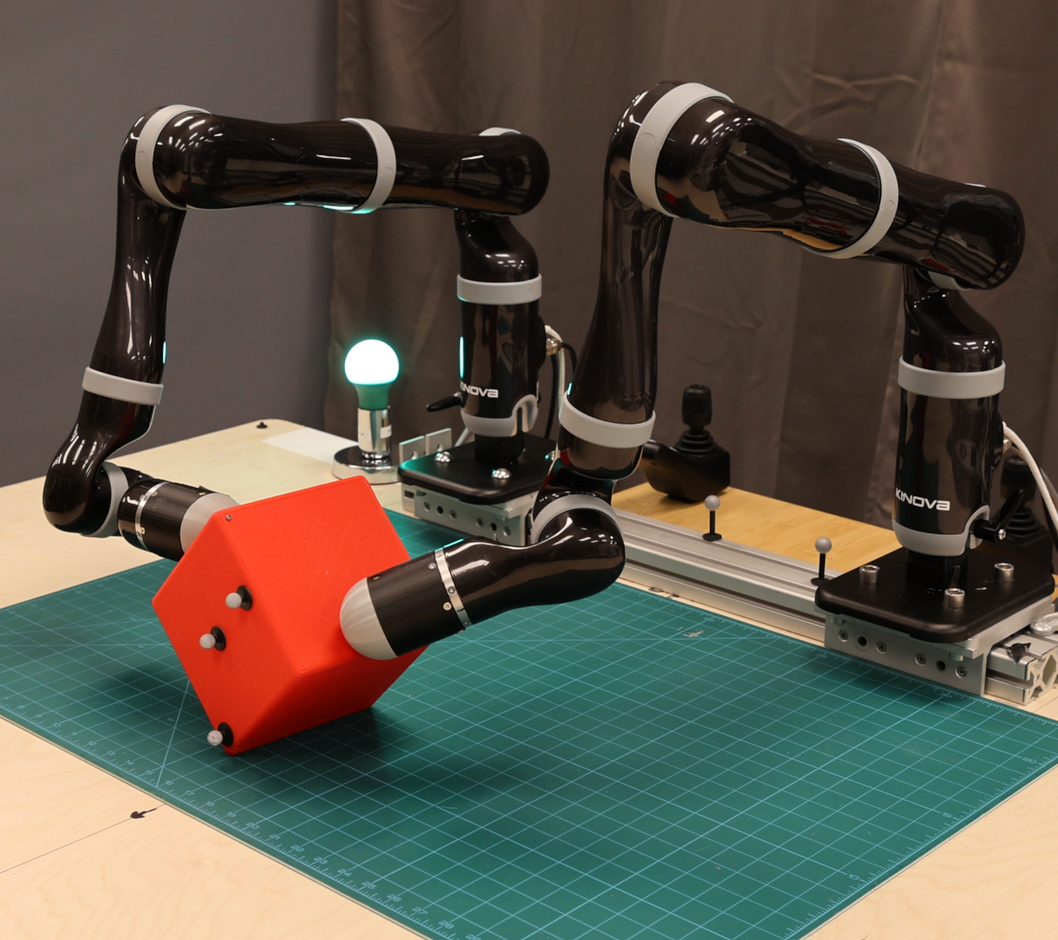}
        \caption{Bi-manual Manip.: 100~Hz}
        \label{fig:cover:dual_jaco}
    \end{subfigure}
    \begin{subfigure}{0.49\linewidth}
        \centering
        \vspace{0.5em}
        \includegraphics[width=0.95\linewidth]{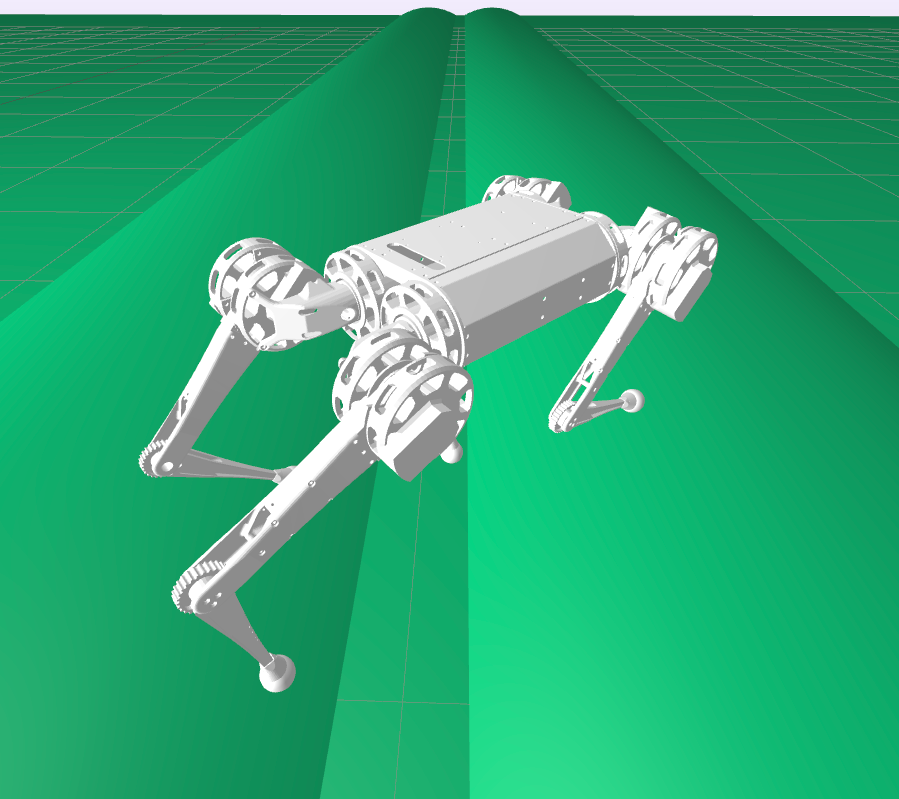}
        \caption{Quadruped: 60~Hz}
        \label{fig:cover:quadruped}
    \end{subfigure}
    \begin{subfigure}{0.49\linewidth}
        \centering
        \vspace{0.5em}
        \includegraphics[width=0.95\linewidth]{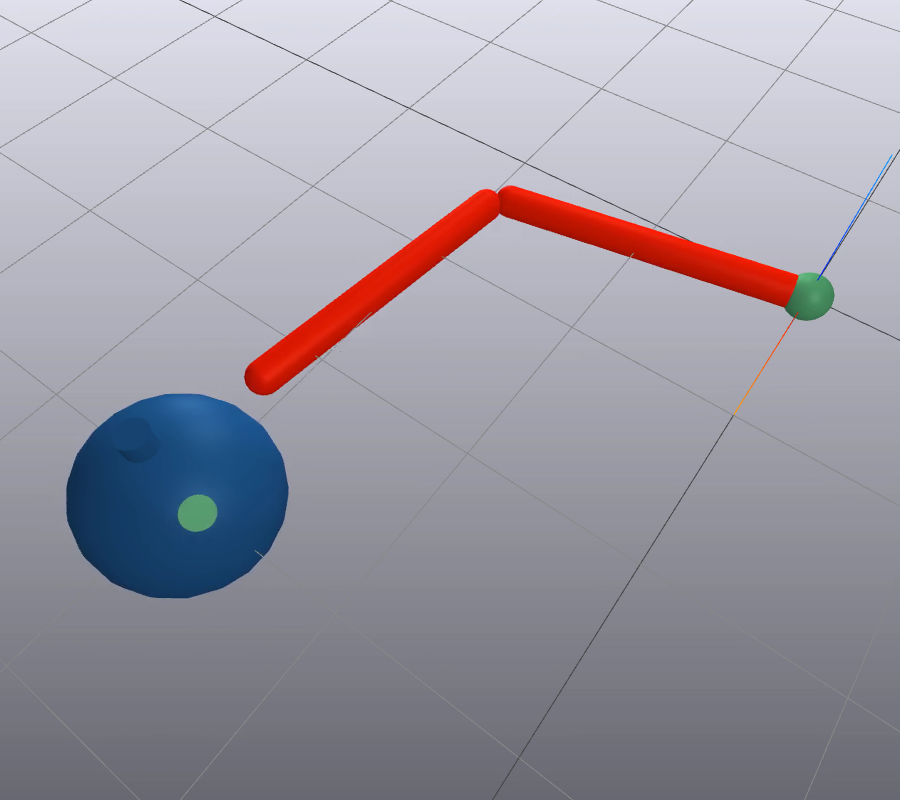}
        \caption{Spinner: 200~Hz}
        \label{fig:cover:spinner}
    \end{subfigure}
    \caption{IDTO enables real-time CI-MPC on a variety of challenging
    manipulation and locomotion tasks. Contact sequences, locations, and timings
    are all determined automatically by the solver over a 1-2 second horizon.}
    \label{fig:cover}
\end{figure}

Here, we adopt a less popular approach, Inverse Dynamics Trajectory Optimization
(IDTO) \citep{erez2012trajectory,todorov2019acceleration}, and show that it
enables fast and effective CI-MPC. Instead of constraints, IDTO uses compliance
and regularized friction to formulate an optimization problem where generalized
positions are the only decision variables. We cast this optimization as a least
squares problem, for which we develop a custom trust-region solver. This solver
leverages a series of small innovations---a smooth and compliant contact model,
scaling, sparse Hessian factorization, and an equality-constrained dogleg
method---to achieve state-of-the-art performance. Despite using a relaxed 
contact model, our IDTO solver enables high-performance real-time CI-MPC on 
systems with rigid contact. We demonstrate this performance on a variety of
simulation examples (where the simulator uses a rigid contact model) as well as
on hardware, as shown in Fig.~\ref{fig:cover}.

\section{Related Work}
\label{sec:related_work}

This section reviews related literature, categorizing methods
based on offline CITO versus online CI-MPC. 

\subsection{Contact-Implicit Trajectory Optimization}

CITO aims to optimize the robot's trajectory, contact forces, and actuation.
Different formulations exist, depending primarily on the choice of decision
variables.

Direct methods introduce decision variables for both state and control, and
enforce the dynamics with constraints \citep{posa2013direct,
manchester2019contact, winkler2018gait, patel2019contact, moura2022non,
wang2023contact}. This results in a large but sparse nonlinear program. Direct
methods support arbitrary constraints and infeasible initializations. However,
the large number of nonlinear constraints used to model dynamics and contact
lead to a large optimization problem, which is difficult to solve efficiently.
Moreover, these formulations can fall into local minima that might not obey
physics \citep{posa2016optimization}.

Shooting methods like Differential Dynamic Programming (DDP)
\citep{mayne1966second} and iterative LQR (iLQR) \citep{li2004iterative} introduce
decision variables only for the control, and are increasingly popular for CITO
\citep{tassa2012synthesis, neunert2017trajectory, carius2018trajectory,
chatzinikolaidis2021trajectory,kim2022contact, howell2022trajectory}. These
methods enforce dynamics with forward rollouts, so each iteration is dynamically
feasible. On the other hand, designing an informative initial guess can be
challenging, and it is difficult to include constraints. Additionally, each
rollout simulates contact dynamics to high precision, which usually requires
solving an optimization sub-problem for each time step
\citep{castro2022unconstrained}. Combining the flexibility of direct methods with
the defect-free integration of shooting is an area of active research
\citep{onol2019contact, onol2020tuning, suh2022bundled, giftthaler2018family}. 

\subsection{Contact-Implicit Model Predictive Control}

The earliest simulated CI-MPC \citep{tassa2012synthesis} ran in real time for
low-dimensional problems, but could not maintain real-time performance for
larger systems. The first hardware implementation \citep{neunert2018whole}
emerged several years later, though this method required extensive parameter
tuning. 

Recent years have seen numerous exciting advancements in CI-MPC. MuJoCo MPC
\citep{howell2022predictive} enables real-time CI-MPC for a number of larger
systems (quadrupeds, humanoids, etc.) using various algorithms.
\cite{cleac2023fast} achieved quadruped CI-MPC on hardware with a bi-level
scheme, performing more expensive computations offline about a reference
trajectory and solving smaller problems online with a sparse interior-point
solver.

\cite{kim2023contact} take a shooting approach, using a mixture of compliant and
rigid contact models. A rigid complementarity-based model is used for forward
rollouts, while gradients are computed based on a smooth relaxation that allows
force at a distance. They demonstrate the performance of this approach in
impressive hardware experiments with a quadruped robot.

\cite{kong2022hybrid} introduce a hybrid-systems iLQR method, with sophisticated
mechanisms for handling unexpected contact mode transitions, and demonstrate
real-time CI-MPC on a quadruped. Another hybrid MPC approach based on the
Alternating Direction Method of Multipliers (ADMM)
\citep{aydinoglu2023consensus} breaks the dependence of the contact-scheduling
problem between time steps, allowing parallelization. The ADMM method enables a
contact-rich ball-rolling task on hardware.

\subsection{Inverse Dynamics Trajectory Optimization}

While most existing work treats inputs as the decision variables (shooting
methods) or states and inputs as the decision variables (most direct methods),
IDTO uses only generalized positions \citep{erez2012trajectory}.

This offers several advantages. Inverse dynamics are faster to compute than the
forward dynamics needed in shooting methods \citep{ferrolho2021inverse}. Unlike
typical direct methods, IDTO does not require complex dynamics constraints, and
uses compliant contact to eliminate the need for contact constraints. Early CITO
work used IDTO and simplified physics to synthesize contact-rich behaviors for
animated characters \citep{mordatch2012contact,mordatch2012discovery}.
\cite{erez2012trajectory} proposed a physically consistent version, and Emo
Todorov described a more mature implementation in Optico, an unreleased software
package, in several talks \citep{todorov2019acceleration, todorov2019optico}. Our
work is heavily inspired by Optico, though we believe our solver differs in a
number of important respects. We use trust region rather than linesearch, for
example, and handle underactuation differently.

A drawback of IDTO is the need for compliant contact, as contact forces must be
a function of state. Much of the community holds complementarity-based rigid
contact as the gold standard for physical realism in planning and simulation
\citep{howell2022dojo, aydinoglu2023consensus, kim2023contact}. Nonetheless, all
contact models are approximations, and strict complimentarity can introduce
non-physical artifacts that are even more severe than those in compliant models
\citep[Section~IV.A]{castro2023theory}. In this paper, we introduce a compliant
contact model with physics-based relaxations, and show that CI-MPC over this
relaxed model performs well on systems with rigid contact. In all of our
simulation experiments, we plan using the relaxed model but simulate using
Drake's state-of-the-art model of rigid contact \citep{drake}. 

\section{Contribution}
\label{sec:novel_contributions}

In this paper, we show that IDTO enables real-time CI-MPC for complex systems
like those shown in Fig.~\ref{fig:cover}. We detail design choices for the
\textit{formulation} as a least-squares problem \hl{and develop a} custom
\textit{solver} for that problem.

Our implementation is available at \texttt{\url{https://idto.github.io}}. In
this paper, we describe implementation details, characterize numerics and
convergence, and evaluate performance. We validate our solver in simulation for
a variety of dexterous manipulation and legged locomotion tasks, as well as on
hardware for bi-manual manipulation. While our solver uses a relaxed contact
model, our simulation experiments use the state-of-the-art physics
in Drake \citep{drake}.

Our approach is not perfect, nor is it the ultimate CI-MPC solution. We strove
to make simple design choices whenever possible, with the hope that this will
illuminate key challenges and pave the way to better solutions in the future. To
this end, we provide an extensive discussion of the limitations of our approach
in Section~\ref{sec:limitations}.

\section{Nonlinear Least-Squares Formulation}
\label{sec:problem_formulation}

We closely follow the notation of our previous work on contact modeling for
simulation \citep{castro2020transition, castro2022unconstrained}. We use $\x = [\q~
\v]$ to describe the state of a multibody system, consisting of generalized
positions $\q\in\mathbb{R}^{n_q}$ and generalized velocities
$\v\in\mathbb{R}^{n_v}$. Time derivatives of the positions relate to
velocities by the kinematic map $\N(\q)\in\mathbb{R}^{n_q\times
n_v}$,
\begin{equation}
    \dot{\q} = \N(\q)\v.
    \label{eq:kinematic_mapping}
\end{equation}
$\N(\q)$ is the identity matrix in most cases, though not for quaternion Degrees of
Freedom (DoFs).

We consider CITO problems of the form
\begin{subequations}\label{eq:continuous_contact_implicit}
\begin{align}
    \min_{\x,\u} ~& \int_{t=0}^T \ell\big(\x(t), \u(t), t\big) dt + \ell_f\big(\x(T)\big),
    \label{eq:continuous_cost}\\
    \mathrm{s.t.~} & \M(\q)\dot{\v} + \C(\q,\v) = \B \u + \J^T \f,
     \label{eq:continuous_dynamics_constraint} \\
    & \x(0) = \x_0.
    \label{eq:initial_condition}
\end{align}
\end{subequations}

The cost \eqref{eq:continuous_cost} encodes the task and penalizes actuation,
where $\ell$ and $\ell_f$ are running and terminal costs respectively. The
problem is constrained to satisfy the equations of motion
\eqref{eq:continuous_dynamics_constraint}, where $\M$ is the mass matrix, $\C$
collects gyroscopic terms, gravitational forces and joint damping, and $\B$ maps
actuation inputs $\u\in\mathbb{R}^{n_u}$ to actuated DoFs. The dynamics
\eqref{eq:continuous_dynamics_constraint} include both the robot and objects in
the environment. For a set of $n_c$ contact constraints, contact forces
$\f\in\mathbb{R}^{3n_c}$ are applied through the contact Jacobian
$\J\in\mathbb{R}^{3n_c\times n_v}$. Finally, \eqref{eq:initial_condition}
provides the initial condition at time $t=0$.

\subsection{Contact Kinematics}
Given a configuration $\q$, our contact engine reports a set of $n_c$ contact
pairs. We characterize the $i\text{-th}$ pair by its location, signed distance
$\phi_i(\q)\in\mathbb{R}$ and contact normal $\widehat{\bm{n}}_i$, see
\citep{castro2022unconstrained} for details. The relative velocity at the contact
point is denoted $\bm{v}_{c, i}$, and relates to generalized velocities via the
$i\text{-th}$ row of the contact Jacobian. We split the contact velocity into
its normal component $v_{n,i} = \widehat{\bm{n}}_i\cdot\bm{v}_{c, i}$ and
tangential component $\bm{v}_{t,i} = \bm{v}_{c,i} - v_{n,i}\widehat{\bm{n}}_i$,
so that $\bm{v}_{c, i}=[v_{n, i}~\bm{v}_{t, i}]$. For brevity, we omit the
contact subscript $i$ from now on.

\subsection{Contact Modeling}
\label{sec:formulation:contact}

Many CITO formulations augment \eqref{eq:continuous_contact_implicit} with
additional decision variables and constraints to model contact forces that
satisfy Coulomb friction and the principle of maximum dissipation. Building on
our experience in contact modeling for simulation \citep{castro2020transition,
castro2022unconstrained}, we adopt a simpler approach based on compliant contact
with regularized friction. With this approach, contact forces are a function of
state. Unlike the accurate and often stiff models used for simulation, however,
here we focus on algebraic forms that are continuously differentiable and thus
more suitable for trajectory optimization. As we will show in
Section~\ref{sec:results}, this relaxed contact model enables performant CI-MPC
in simulations with a more realistic model of rigid contact as well as on
hardware.

\begin{figure*}
    \centering
    \begin{subfigure}{0.3\linewidth}
        \centering
        \includegraphics[width=\linewidth]{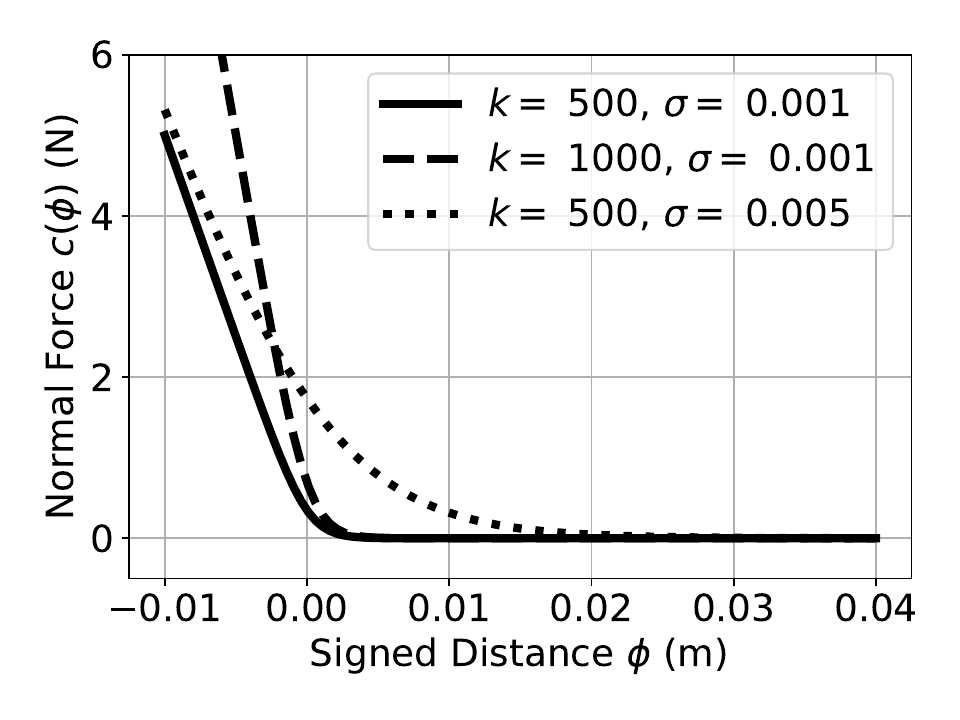}
        \caption{\hl{Stiffness component $c(\phi)$ \eqref{eq:compliance_model}.}}
        \label{fig:stiffness}
    \end{subfigure}
    \begin{subfigure}{0.3\linewidth}
        \centering
        \includegraphics[width=\linewidth]{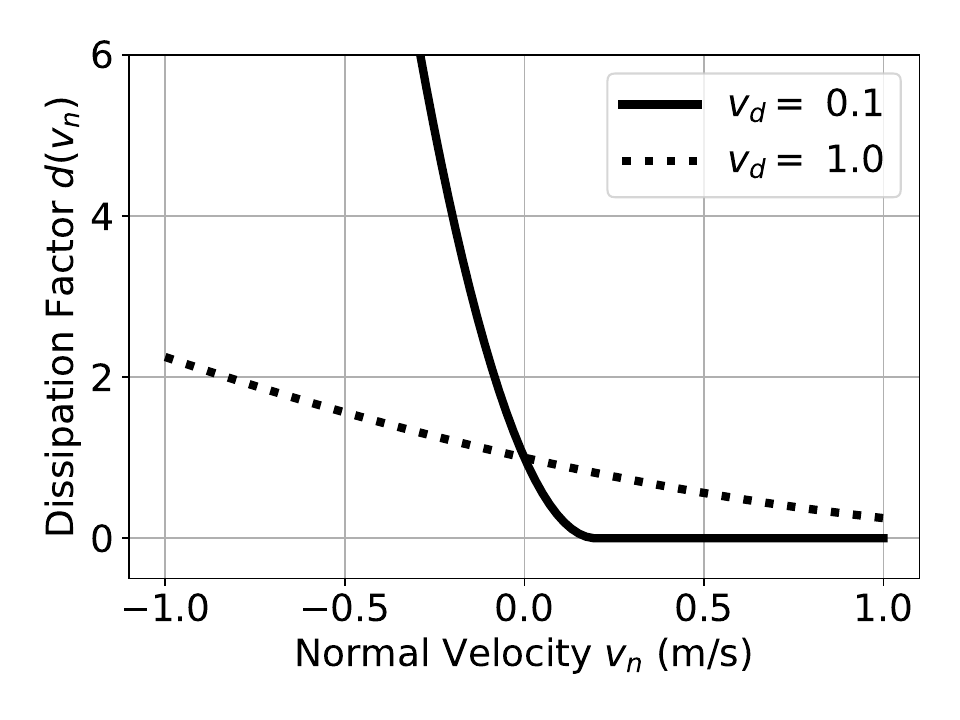}
        \caption{\hl{Dissipation component $d(v_n)$ \eqref{eq:dissipation_model}.}}
        \label{fig:dissipation}
    \end{subfigure}
    \begin{subfigure}{0.3\linewidth}
        \centering
        \includegraphics[width=\linewidth]{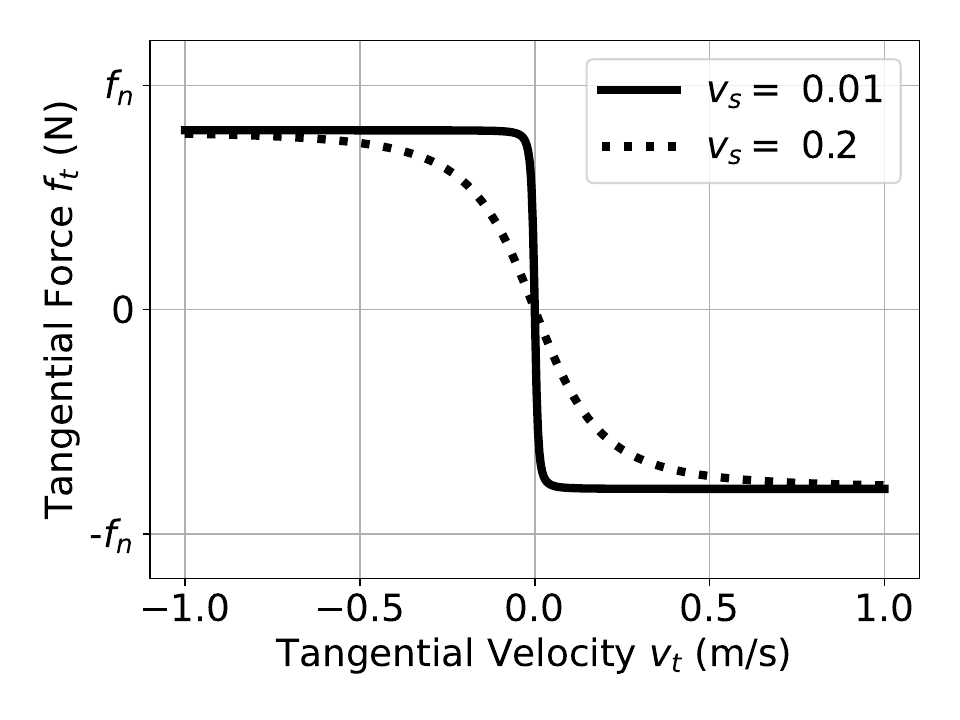}
        \caption{\hl{Regularized friction \eqref{eq:regularized_friction_model} with $\mu = 0.8$.}}
        \label{fig:friction}
    \end{subfigure}
    \caption{\hl{Visualization of our compliant contact model
    \eqref{eq:compliance_model}-\eqref{eq:regularized_friction_model} for
    various parameter values. This model is differentiable everywhere, but
    introduces force-at-a-distance (\subref{fig:stiffness}) and
    drift-during-stiction (\subref{fig:friction}) artifacts.}}
    \label{fig:contact_model}
\end{figure*}

As with the contact velocity, we split the contact force at each contact point
into its normal $f_n$ and tangential $\bm{f}_t$ components such that
$\bm{f}_{c}=[f_{n}~\bm{f}_{t}]$. \hl{The key aspects of our contact model are
illustrated in Fig.~\ref{fig:contact_model}, and detailed below.}

We model the normal force as
\begin{equation}
    f_n = c(\phi)d(v_n),
\end{equation}
where
\begin{equation}\label{eq:compliance_model}
    c(\phi) = \sigma k \log\left(1 + \exp(-\phi/ \sigma)\right)
\end{equation}
provides a smooth model of compliance that approximates a linear spring of
stiffness $k$ in the limit to zero smoothing parameter $\sigma$
\hl{(Fig.~\ref{fig:stiffness})}. While $\sigma$ can be thought of as the scale
parameter of a logistic distribution over signed distances $\phi$
\citep{pang2022global}, here we use it to trade off smoothness and force at a
distance. \hl{$\sigma$ takes length units, and should be set according to the
scale of the system in question.}

We model dissipation as
\begin{equation}\label{eq:dissipation_model}
    d(v_n) = 
    \begin{cases}
        1 - \frac{v_n}{v_d} & \text{if } \frac{v_n}{v_d} < 0, \\
        (\frac{v_n}{v_d} - 2)^2 / 4 & \text{if }  0 \leq \frac{v_n}{v_d} < 2, \\
        0 & \text{if } 2 \leq \frac{v_n}{v_d},
    \end{cases}
\end{equation}
a smoothed Hunt and Crossley model \citep{hunt1975coefficient} with dissipation
velocity $v_d = 1/d_{HC}$ the reciprocal of the Hunt and Crossley dissipation
parameter $d_{HC}$ \hl{(Fig.~\ref{fig:dissipation}). The intuition behind this
model is as follows: if two bodies are moving toward each other, $d(v_n)$
increases the effective stiffness of the contact. Conversely, if the bodies are
moving away from each other ($v_n > 0$), the effective stiffness is reduced.
This helps avoid large interpenetrations while also limiting the ``bounciness''
of the contact.}

We model the tangential component $\f_t$ with a regularized model of Coulomb
friction
\begin{equation}\label{eq:regularized_friction_model}
    \f_t = - \mu \frac{\v_t}{\sqrt{v_s^2 + \|\v_t\|^2}}\,f_n,
\end{equation}
where $\mu$ is the friction coefficient and $v_s$ is the \emph{stiction
tolerance} \hl{(Fig.~\ref{fig:friction})}. This model satisfies both the model of Coulomb friction
($\Vert\bm{f}_t\Vert\le \mu f_n$) and the principle of maximum dissipation
(friction opposes velocity) at the expense of drift during stiction at
speeds lower than $v_s$.

\hl{

\subsubsection{Choosing Contact Parameters}\label{sec:formulation:contact:params}

Unlike contact modeling for simulation, where the principal objective is to
model physical dynamics as faithfully as possible, contact modeling for CI-MPC
has two competing goals: (1) model the system accurately, and (2) maintain a
friendly optimization landscape. Modeling choices that improve physical
accuracy, like high stiffnesses and tight friction regularization, also make the
optimization problem more difficult, inducing narrow valleys and steep walls in
the cost landscape.

As we will show in Section~\ref{sec:results}, CI-MPC feedback is able to make up
for significant modeling errors. As a result, we recommend choosing contact
parameters \textit{as soft/highly regularized as possible, while maintaining
good closed-loop performance with a rigid simulator.}
Table~\ref{tab:contact_params} lists each of the parameters in our contact
model, along with its physical meaning and our tuning recommendations.

While contact parameter tuning is an ad-hoc and problem-specific process, the
rapid feedback enabled by a fast solver like IDTO does make parameter tuning
easier. This could be further improved with a graphical interface that allows
for tuning in real-time, as in \cite{howell2022predictive}. Additionally, each
of the contact parameters has a clear physical interpretation, which can guide
the tuning process. This is not the case for all common contact models, as
illustrated in \cite[Section IV]{castro2023theory}.

\begin{table*}
    \centering
    \caption{\hl{Summary of contact parameters and tuning recommendations.}}
    \hl{
    \begin{tabular}{ccp{6cm}p{8cm}}
        \toprule
        \textbf{Symbol} & \textbf{Units} & \textbf{Physical meaning} & \textbf{Tuning recommendation} \\
        \midrule
        $k$ & N/m & Spring stiffness for contact normal force. & Set as low as possible while maintaining good closed-loop MPC performance with a rigid simulator. \\
        \midrule
        $\sigma$ & m & Amount of force-at-a-distance/smoothing factor. & Scale with the size of the system in question, set as large as possible while maintaining closed-loop performance. \\
        \midrule
        $v_d$ & m/s & Dissipation velocity: lower values correspond to stiffer and less bouncy contact. & $0.1$ m/s worked well for all of the systems we tried. \\
        \midrule
        $\mu$ & - & Friction coefficient. & Set according to the physical system in question. \\
        \midrule
        $v_s$ & m/s & Stiction velocity: the speed at which bodies in stiction slide. & Set as high as possible while maintaining good closed-loop MPC performance with a rigid simulator. \\
        \midrule
    \end{tabular}
    }
    \label{tab:contact_params}
\end{table*}
}

\subsection{Inverse Dynamics Trajectory
Optimization}\label{sec:formulation:idto}

In this section, we approximate (\ref{eq:continuous_contact_implicit}) as a
nonlinear least squares problem. The key idea is to use generalized positions
$\q$ as the only decision variables and enforce dynamic feasibility with inverse
dynamics. 

We will focus on quadratic cost terms of the form
\begin{gather}
    \ell(\x,\u,t) = \frac{1}{2}\|\x(t)-\bar{\x}(t)\|^2_{\Q} + \frac{1}{2}\|\u(t)\|^2_{\R}, \nonumber \\
    \ell_f(\x) = \frac{1}{2}\|\x(T)-\bar{\x}(T)\|^2_{\Q_f},
    \label{eq:quadratic_cost}
\end{gather}
where $\Q$, $\R$, and $\Q_f$ are diagonal scaling matrices and $\bar{\x}(\cdot)$
is a nominal trajectory. $\bar{\x}(\cdot)$ outlines a desired behavior and is
not necessarily dynamically feasible. 

We discretize the time horizon into $N$ steps of size $\delta t$ and approximate
\eqref{eq:continuous_cost} using a first-order quadrature rule
\begin{equation}\label{eq:discrete_cost}
    L(\x,\u) = \sum_{k=0}^{N} \ell_k(\x_k, \u_k),
\end{equation}
where for convenience we define the total cost $L(\x,\u)$, $\ell_k = \delta
t\ell(\x_k, \u_k)$, and $\ell_N = \ell_f(\x_N)$.

We then proceed to write (\ref{eq:discrete_cost}) in terms of generalized
positions $\q$.\footnote{We denote the vector of all configurations with
$\q = [\q_0, \q_1, \dots, \q_N]$ and similarly all velocities $\v =
[ \v_0, \v_1, \dots, \v_N]$. } Velocities $\v$ are a function of $\q$ as
follows,
\begin{equation}
    \v_{k}(\q) = \N^+(\q_k)\frac{\q_k - \q_{k-1}}{\delta t}\quad\forall k=1\dots N, 
\end{equation}
where $\N^+(\q_k)$ is the left pseudoinverse of the kinematic mapping matrix 
\eqref{eq:kinematic_mapping}. $\v_{0} = \v(0)$ is given by
\eqref{eq:initial_condition}.

Similarly, we approximate accelerations $\a$ as
\begin{equation}
    \a_k(\q) = \frac{\v_{k+1}(\q) - \v_k(\q)}{\delta t}\quad\forall k=0\dots N-1.
\end{equation}
Note that we use $k+1$ and $k$ for accelerations, while we use $k$ and $k-1$ for
velocities. This leads to $\a_k$ that depend symmetrically on $\q_{k-1}$,
$\q_k$, and $\q_{k+1}$.

We write contact forces $\f$ as a function of $\q$ using the compliant contact
model outlined in Section~\ref{sec:formulation:contact},
\begin{equation}
    \f_k(\q) = \f(\q_{k+1}, \v_{k+1}(\q))\quad\forall k=0\dots N-1.
\end{equation}

Finally, we use inverse dynamics to define the generalized forces $\btau_k$
needed to advance the state from time step $k$ to time step $k+1$,
\begin{multline}
    \btau_k(\q) = \M(\q_{k+1})\a_k + \C(\q_{k+1}, \v_{k+1}) \\
     -\J^T(\q_{k+1}) \f_k(\q_{k+1}, \v_{k+1}).
    \label{eq:discrete_inverse_dynamics}
\end{multline}

When compared to a time-stepping scheme for simulation
\citep{castro2022unconstrained}, all terms are evaluated \emph{implicitly}. We
make this choice based on the intuition that implicit time-stepping schemes are
stable even for stiff systems of equations. \hl{The accuracy and usefulness of
this intuition in the IDTO setting is an important area for future study.}

For fully actuated systems, generalized forces equal control torques, $\btau_k =
\u_k$. In the more typical underactuated case, we have $\btau_k = \B \u_k$, and
some entries of $\btau_k$ must be zero. Substituting both $\btau_k$ and $\x_k = [\q_k~\v_k]$ as functions of $\q$ in
\eqref{eq:discrete_cost}, we can approximate
(\ref{eq:continuous_contact_implicit}) as,
\begin{subequations}\label{eq:id_traj_opt}
    \begin{align}
        \min_{\q}~     & L(\q),
        \label{eq:discrete_total_cost} \\
        \mathrm{s.t.~} & \h(\q) = 0,
        \label{eq:underactuation_constraint}
    \end{align}
\end{subequations}
where $\h(\q)$ collects unactuated rows of $\btau_k(\q)$.

\begin{remark}\label{remark:least_squares} This is a nonlinear least squares
    problem with equality constraints. To see this, define the residual
    \begin{equation}
        \bm{r}(\q) = 
        \begin{bmatrix}
            \bm{\tilde{Q}}^{1/2}(\x(\q)-\bar{\x}) \\
            \bm{\tilde{R}}^{1/2}\btau(\q)
        \end{bmatrix}
        \label{eq:ls_residual}
    \end{equation}
    where $\bm{\tilde{Q}}$ and $\bm{\tilde{\R}}$ denote the diagonal matrices
    that result from stacking $\Q$ and $\R$. We can then write the cost in standard least squares form: $L(\q) =
    1/2\Vert\bm{r}(\q)\Vert^2$.
\end{remark}

\section{Gauss-Newton Trust-Region Solver}
\label{sec:solver}

In this section, we develop a solver tailored to \eqref{eq:id_traj_opt}. 

\subsection{The Unconstrained Problem}
\label{sec:solver:dogleg}

\begin{figure}
    \centering
    \includegraphics[width=0.85\linewidth]{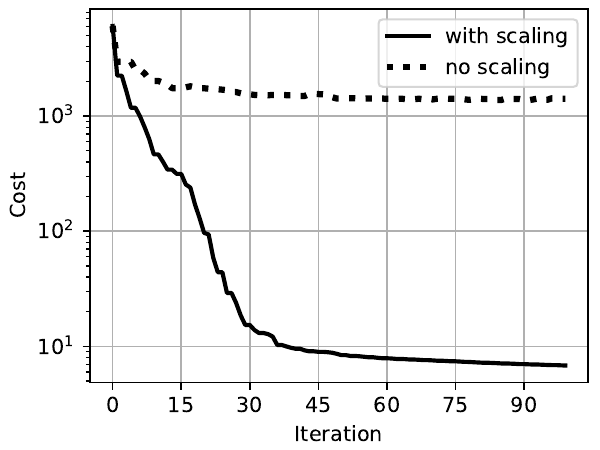}
    \caption{Convergence plot for the Allegro hand example with and without scaling.}
    \label{fig:scaling_comparison}
\end{figure}

We start by considering the unconstrained problem, for which we choose a
trust-region method. At each iteration $i$, this method finds an update
$\q^{i+1} =
\q^i + \p$ by solving a quadratic approximation of
\eqref{eq:id_traj_opt} around $\q^i$,
\begin{subequations}
    \label{eq:tr_subproblem}
\begin{align}
    \min_{\p} ~& L^i + \g^T\p + \frac{1}{2}\p^T\H\,\p
    \label{eq:tr_subproblem_cost}\\
    \mathrm{s.t.} ~& \|\D^{-1}\p\| \leq \Delta,
    \label{eq:tr_constraint}
\end{align}
\end{subequations}
where $L^i=L(\q^i)$ and $\g = \nabla L(\q^i)$ are the cost and its gradient.
Matrix $\H$ is a Gauss-Newton approximation of the Hessian, see Section
\ref{sec:solver:gauss_newton} for details. At each iteration, the trust-region
constraint \eqref{eq:tr_constraint} limits the size of the step $\p$. We use a
diagonal matrix $\D$ to scale the problem and improve numerical conditioning.
Stiff contact dynamics result in a cost landscape with narrow valleys and steep
walls. This makes scaling critical for good performance, as illustrated in
Fig.~\ref{fig:scaling_comparison}. We tried various strategies
\citep{more2006levenberg}, and found that $\mathrm{diag}(\D) =
\mathrm{diag}(\H)^{-1/4}$ gave the best results. 

Since solving \eqref{eq:tr_subproblem} exactly would be computationally
expensive, we use the \emph{dogleg method} \cite[\S 4.1]{nocedal1999numerical}
to find an approximate solution. This requires a single update and factorization
of $\H$ per iteration. After each iteration, we update the trust-region radius
$\Delta$ with a standard strategy \cite[Algorithm~4.1]{nocedal1999numerical}.

\begin{remark}\label{remark:tr_vs_linesearch} We chose trust region over
    linesearch for several reasons. One is speed: each evaluation of the cost
    $L(\q)$ is expensive, and trust region requires us to compute $L(\q)$ only
    once per iteration. Trust region is also appealing because $\H$ is poorly
    conditioned. In some extreme instances, this can lead to Gauss-Newton search
    directions ($\p = -\H^{-1}\g$) that are not actually descent directions due
    to floating point error. Trust-region methods turn toward gradient descent
    as the trust radius is reduced, which helps avoid this issue.   
\end{remark}

\subsection{Handling Equality
Constraints}\label{sec:solver:equality_constraints}

For systems of practical interest with unactuated DoFs, enforcing
\eqref{eq:underactuation_constraint} is critical for dynamic feasibility. We
consider two methods to handle this constraint.

\subsubsection{Quadratic Penalty}\label{sec:solver:penalty}

This method adds a penalty of the form $\frac{w}{2}\Vert\h(\q)\Vert^2$ to the cost
$L(\q)$, where $w$ is a large scalar weight. In practice we implement this by
placing $w$ at elements of $\R$ that correspond to unactuated DoFs. 

\subsubsection{Lagrange Multipliers}\label{sec:solver:lagrange}

The penalty method is simple to implement, and worked well in many cases.
However, it cannot entirely drive unactuated torques to zero without high
penalties that worsen numerics.

As an alternative, we developed a constrained dogleg method based on Lagrange
multipliers. We start with a constrained Newton step
\begin{equation}\label{eq:constrained_newton_step}
    \begin{bmatrix}
        \H & \A^T \\
        \A & \bm{0}
    \end{bmatrix}
    \begin{bmatrix}
        \p \\
        \blambda
    \end{bmatrix} = 
    \begin{bmatrix}
        -\g \\ -\h
    \end{bmatrix},
\end{equation}
where $\A$ is the Jacobian of $\h$ and $\blambda$ is a vector of Lagrange
multipliers. Since our Hessian approximation is invertible, we can
eliminate $\p$ from \eqref{eq:constrained_newton_step} and solve for
\begin{equation}
    \label{eq:lagrange_multiplier}
    \blambda = (\A\,\H^{-1} \A^T)^{-1}(\h - \A \H^{-1} \g).
\end{equation}
We then solve \eqref{eq:constrained_newton_step} for the Newton step
\begin{equation}
    \p = -\H^{-1}(\A^T\blambda + \g).
\end{equation}
We can view this step as the unconstrained minimizer of a quadratic model of the
merit function 
\begin{equation}\label{eq:al_merit_function}
    L(\q) + \h(\q)^T\blambda.
\end{equation}

This observation motivates a trust-region variant, where we minimize a quadratic
model of (\ref{eq:al_merit_function}) subject to bounds on the step size. More
formally, the new trust-region subproblem is given by
\begin{subequations}
\begin{align}
    \min_{\p} ~& L + (\g + \A^T\blambda)^T\p + \frac{1}{2}\p^T\H\p \\
    \mathrm{s.t.} ~& \|\D^{-1}\p\| \leq \Delta,
\end{align}
\end{subequations}
which we can solve with the dogleg method. This approach is similar to
Fletcher's smooth exact penalty method \citep{nocedal1999numerical}, albeit with
a slightly different approximation of the Lagrange multipliers $\blambda$.

Computing $\blambda$ efficiently does take some care. We use our sparse
factorization of $\H$ to compute $\H^{-1}\A^T$ one column at a time. We then
form a dense $\A\H^{-1}\A^T$ matrix and solve for $\blambda$ with a dense
Cholesky factorization. It should be possible to exploit the sparse structure of
$\A\H^{-1}\A^T$ to perform this inversion more efficiently, but we have not done
so for this work.

\subsection{Sparse Gauss-Newton Hessian Approximation}
\label{sec:solver:gauss_newton}

Note that $\ell_{k}$ does not depend on the full configuration sequence $\q$ but
only on the stencil $\q_{k-1}$, $\q_{k}$, $\q_{k+1}$, 
\begin{equation}
    \ell_{k}(\x_k(\q),\btau_k(\q)) = \ell_{k}(\q_{k-1},\q_{k},\q_{k+1}).
\end{equation}

Therefore the $k\text{-th}$ segment of the gradient, $\g_k$, can be computed as
\begin{align}
    \frac{\partial L(\q)}{\partial \q_k} &=\g_k(\q_{k-2},\q_{k-1},\q_{k},\q_{k+1},\q_{k+2})\nonumber\\
        &=\frac{\partial\ell_{k-1}}{\partial \q_k} +
        \frac{\partial\ell_k}{\partial \q_k} +
        \frac{\partial\ell_{k+1}}{\partial \q_k}.
        \label{eq:gt}        
\end{align}
Using the chain rule through the quadratic costs in \eqref{eq:quadratic_cost},
we can write the terms in \eqref{eq:gt} as
\begin{equation}
    \frac{\partial\ell_k}{\partial \q_s} = (\x_k-\bar{\x}_k)^T\Q\frac{\partial \x_k}{\partial \q_s} + \btau_{k}^T \R \frac{\partial \btau_{k}}{\partial \q_s},
    \label{eq:explicit_cost_gradient}
\end{equation}
for $s=k-1\dots k+1$.

Given that the gradient in \eqref{eq:gt} involves a stencil with five time
steps, the Hessian inherits a pentadiagonal sparsity structure
\begin{equation}
    \H_{k,s} = \frac{\partial \g_k}{\partial \q_s}\quad\forall s=k-2 \dots k+2.
\end{equation}
We exploit this structure by factorizing $\H$ with a block-sparse
variation of the Thomas algorithm \citep{benkert2007efficient}.

While it is possible to compute the first and second derivatives of inverse dynamics
analytically \citep{carpentier2018analytical,singh2022efficient}, obtaining
analytical derivatives through contact is significantly more challenging, given
the complex kinematics involved \citep{cui2010geometric}. This motivates a
Gauss-Newton approximation. Applying the chain rule through
\eqref{eq:explicit_cost_gradient} and neglecting second-order derivatives leads
to an approximation in terms of generalized velocity and inverse dynamics
gradients,
\begin{equation}
    \H(\q) \approx \H\left(\frac{\partial \v}{\partial \q}, \frac{\partial \btau}{\partial \q}\right),
\end{equation}
with the full expression given in Appendix~\ref{apx:gauss_newton_hessian}.

We approximate the velocity gradients as
\begin{align*}
    & \frac{\partial \v_k}{\partial \q_k} = \frac{1}{\delta t} \N^+(\q_k^i), \\
    & \frac{\partial \v_{k+1}}{\partial \q_k} = -\frac{1}{\delta t} \N^+(\q_{k+1}^i),
\end{align*}
where we have \emph{frozen} $\N^+$ at the previous iteration $i$.

For the inverse dynamics gradients, we use finite differences. This is
computationally intensive since it requires many inverse dynamics evaluations
and geometry queries. These derivatives are the most time-consuming portion of
IDTO, as shown by the profiling in Section~\ref{sec:results}.

\begin{remark}\label{remark:collision_geometries} Computing gradients through
    contact requires a contact engine at least as accurate as the finite
    difference step size. We found that \hl{geometry computations} based on
    iterative methods \citep{gilbert1988fast} do not provide the required
    accuracy and lead to inaccurate gradients. Furthermore, geometries with
    sharp corners (e.g., boxes) induce discontinuous jumps in the contact normal
    $\widehat{\bm{n}}$, rendering finite difference derivatives useless. We
    therefore restricted ourselves to analytical \hl{geometry computations}
    accurate to machine precision, and replaced collision geometries with
    inscribed spheres or half-spaces.
\end{remark}

\hl{While analytical gradients could greatly improve solver speed and accuracy,
computing them efficiently remains a significant challenge. To the best of our
knowledge, existing analytical methods like \cite{carpentier2019pinocchio} do
not account for the variations in contact \textit{location}. In particular, the
contact jacobian $\J$ depends on the location of contact points. Contact
location then depends on collision geometries and varies in a way that is not
captured by analytical methods designed for trajectory optimization with fixed
contact sequences. For this reason, our solver relies on a simple
finite-difference approach.}

\section{Results}
\label{sec:results}

Here we characterize the performance of our IDTO solver on the four systems
shown in Fig.~\ref{fig:cover}. In all simulation examples, the simulator uses
Drake's state-of-the-art model of rigid contact, while the planner uses the
compliant model described in Section~\ref{sec:formulation:contact}.

\subsection{Test Cases}
\label{sec:example_systems}

\subsubsection{Spinner}

The spinner is a 3-DoF system shown in Fig.~\ref{fig:cover:spinner}. The two
finger links are 1~m long and mass 1~kg, with a diameter of 0.05~m. The spinner
itself is a 1~kg sphere with a 0.25~m radius. The target trajectory $\bar{\x}$
rotates the spinner 2 radians while the finger remains stationary at the initial
position, with the finger tip 0.08~m from the spinner.

\subsubsection{Mini Cheetah Quadruped}

Mini Cheetah (Fig.~\ref{fig:cover:quadruped}) is a 9~kg quadruped with 18 DoFs
(12 actuated joints and a floating base) \citep{katz2019mini}. The robot is
tasked with moving in a desired direction and matching a desired orientation. We
model hills as 1~m diameter cylinders embedded in the ground.

Unlike most existing work on CI-MPC for quadrupeds
\citep{winkler2018gait,cleac2023fast,kong2022hybrid,howell2022predictive}, we do
not specify a preferred gait sequence. Rather, the robot is merely asked to move
in the desired direction, with a small terminal cost encouraging the legs to end
the trajectory in a standing configuration. 

\subsubsection{Allegro Dexterous Hand}

Our highest-DoF example is a dexterous manipulation task with the simulated
Allegro hand shown in Fig.~\ref{fig:cover:allegro}. The ball is a 50~g sphere
with a 6~cm radius. The overall system has 22 DoFs---16 actuated in the
hand and 6 unactuated for the ball.

The hand is to rotate the ball so that the colorful marks on the ball line up
with a target set of axes, shown as a transparent triplet in
Fig.~\ref{fig:cover:allegro} and the accompanying video. The nominal trajectory
$\bar{\x}$ consists of the ball rotating while the hand remains stationary in a
nominal ``holding'' configuration.

\subsubsection{Bi-Manual Manipulator}

\begin{figure}
    \centering
    \includegraphics[width=0.48\linewidth]{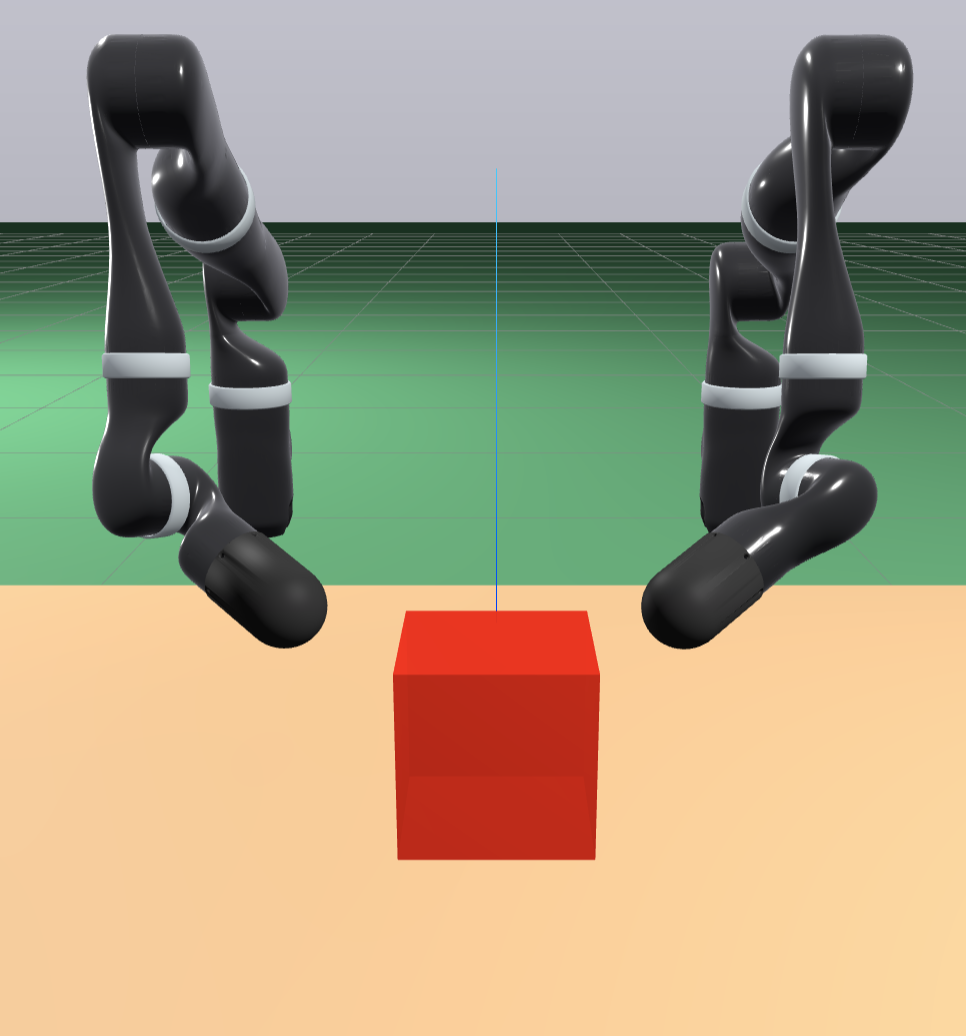}
    \includegraphics[width=0.48\linewidth]{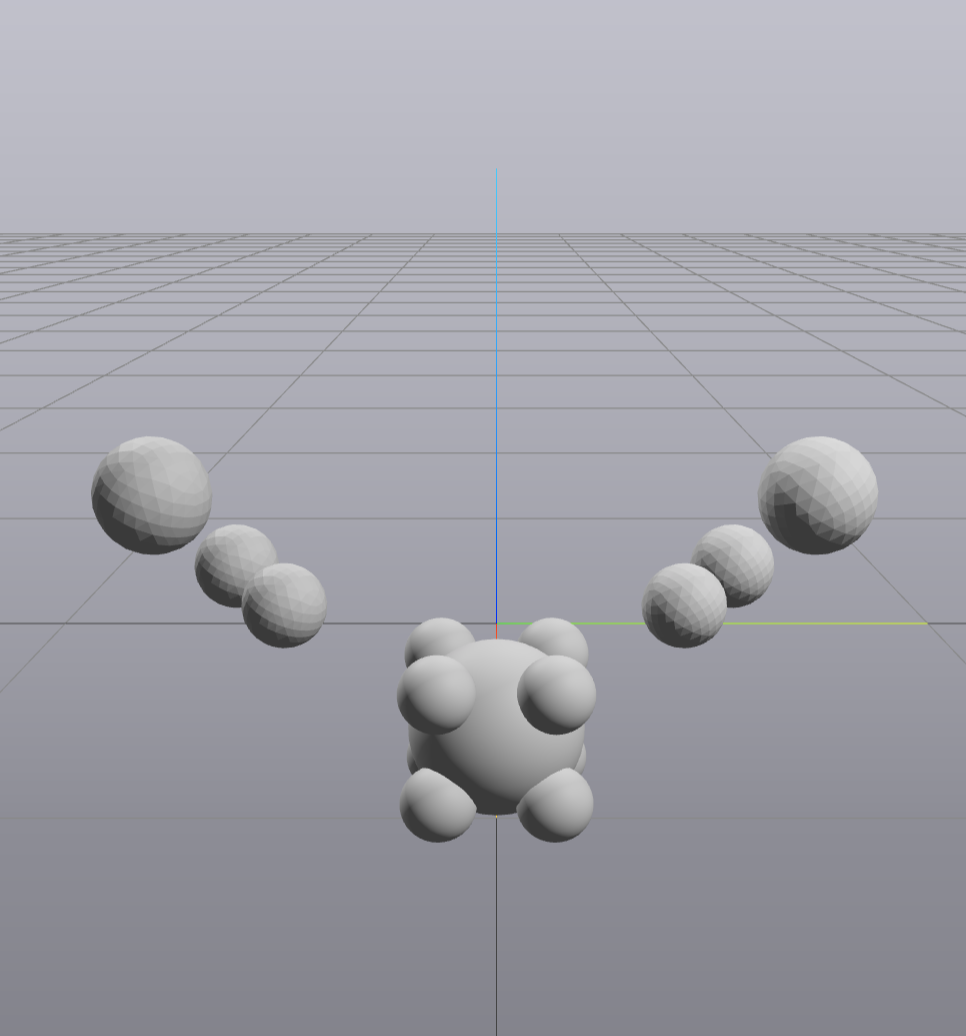}
    \caption{\hl{Visual (left) and collision (right) geometries for the
    bi-manual manipulation task. We eliminate sharp corners by modeling the box
    with inscribed spheres. Additionally, we only include arm collision geometries
    near the end-effectors. IDTO
    is able to overcome the resulting modeling error in both simulations (using
    box collisions) and hardware experiments. }}
    \label{fig:inscribed_spheres}
\end{figure}

Finally, we consider the bi-manual manipulation task shown in
Fig.~\ref{fig:cover:dual_jaco}. This consists of two Kinova Jaco arms and a
550~g uniform density cube with 15~cm sides. The system has 20 DoFs: 14 actuated
joints in the arms plus the box's floating base. As discussed in
Remark~\ref{remark:collision_geometries}, finite differences impose smoothness
and accuracy requirements on the geometry queries, so we model the box
with 9 inscribed spheres, \hl{as shown in Fig.~\ref{fig:inscribed_spheres}}.

The target trajectory $\bar{\x}$ moves the box to a desired pose while keeping
the arms stationary in their initial configuration. We consider three tasks,
each defined by a different target pose: moving the box along the table, lifting
the box up, and balancing the box on its edge. 

\subsection{Parameters}

Table~\ref{tab:contact_parameters} reports contact and planning parameters for
each example system. Note that these contact parameters are used by the
optimizer for planning, but not by the simulator.

Contact stiffness scales with the expected magnitude of contact forces, as well
as the amount of penetration we are willing to accept. The smoothing factor is
determined primarily by the size of the system, with smaller systems requiring a
smaller smoothing length scale. Numerical considerations dominate the choice of
stiction velocity: small values model stiction more accurately, but large values
result in a smoother problem for the optimizer. 

\begin{table}
    \small\sf\centering
    \caption{Contact model and planning parameters.}
    \label{tab:contact_parameters}
    {\footnotesize
    \begin{tabular}{ccccc}
        \toprule
                        & Spinner & Quadruped & Allegro & Bi-Manual \\
        \midrule
        Stiffness (N/m) & 200     & 2000      & 100     & 1000      \\
        Smoothing (cm)   & 1.0    & 1.0      & 0.1   & 0.5     \\
        Friction Coeff. & 0.5     & 1.0       & 1.0     & 0.2       \\
        Stiction (m/s)  & 0.05    & 0.5       & 0.1     & 0.05      \\
        Time Step (s)   & 0.05    & 0.05      & 0.05    & 0.05      \\
        Horizon (s)     & 2.0     & 1.0       & 2.0     & 1.0 \\
        \bottomrule
    \end{tabular}}
\end{table}

\begin{table*}[h]
    \small\sf\centering
    \caption{Cost weights for each example system.}
    \label{tab:cost_weights}
    \begin{tabular}{cccc|ccc|ccc|ccc}
        \toprule
                                     & \multicolumn{3}{c}{Spinner}    & \multicolumn{3}{c}{Quadruped} &
        \multicolumn{3}{c}{Allegro} & \multicolumn{3}{c}{Bi-Manual}                                                                                                                             \\
                                     & $\Q$                           & $\R$                           & $\Q_f$ & $\Q$                & $\R$ & $\Q_f$ & $\Q$  & $\R$ & $\Q_f$ &
        $\Q$                         & $\R$                           & $\Q_f$                                                                                                                    \\
        \midrule
        actuated positions           & 1                              & -                              & 10     & 0                   & -    & 1      & 0.01  & -    & 1      & 0   & -     & 0.1 \\
        unactuated positions         & 1                              & -                              & 10     & 10 (pos.), 1 (ori.) & -    & 10     & 10    & -    & 100    & 10  & -     & 10  \\
        actuated velocities          & 0.1                            & -                              & 0.1    & 0.1                 & -    & 0.1    & 0.001 & -    & 10     & 0.1 & -     & 1   \\
        unactuated velocities        & 0.1                            & -                              & 0.1    & 1                   & -    & 1      & 1     & -    & 10     & 0.1 & -     & 1   \\
        actuated torque              & -                              & 0.1                            & -      & -                   & 0.01 & -      & -     & 0.1  & -      & -   & 0.001 & -   \\
        unactuated torque            & -                              & 1000                           & -      & -                   & 100  & -      & -     & 1000 & -      & -   & 1000  & -   \\
        \bottomrule
    \end{tabular}
\end{table*}

For the cost weights ($\Q, \R, \Q_f$), we use diagonal matrices with values
reported in Table~\ref{tab:cost_weights}. The final row specifies a quadratic
penalty on torques applied to unactuated DoFs. Note that for the quadruped, we
apply a different weight for the floating base position and orientation in the
running cost.

\subsection{Open-Loop Trajectory Optimization}\label{sec:open_loop_results}

\begin{figure*}
    \centering
    \begin{subfigure}{0.47\linewidth}
        \centering
        \includegraphics[width=0.9\linewidth]{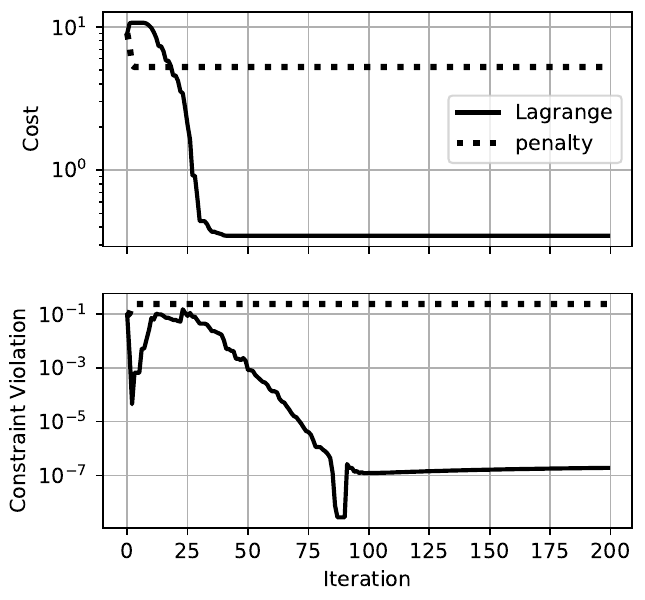}
        \caption{Spinner}
        \label{fig:open_loop_convergence:spinner}
    \end{subfigure}
    \begin{subfigure}{0.47\linewidth}
        \centering
        \includegraphics[width=0.9\linewidth]{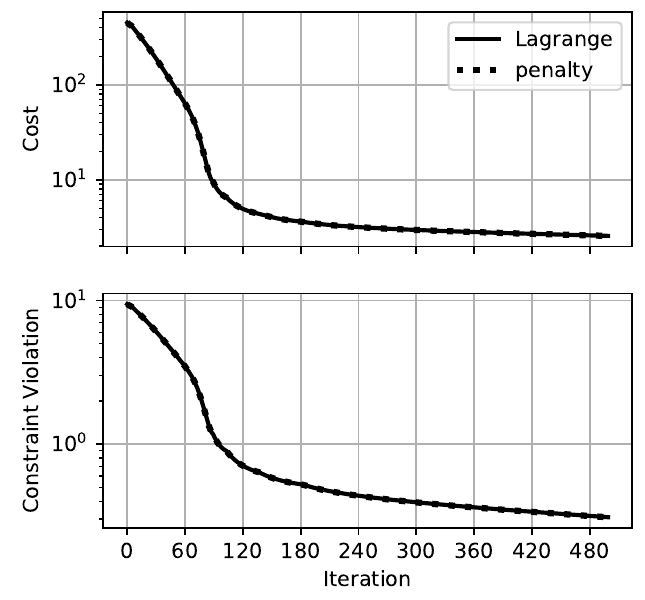}
        \caption{Quadruped}
        \label{fig:open_loop_convergence:quadruped}
    \end{subfigure}
    \begin{subfigure}{0.47\linewidth}
        \centering
        \includegraphics[width=0.9\linewidth]{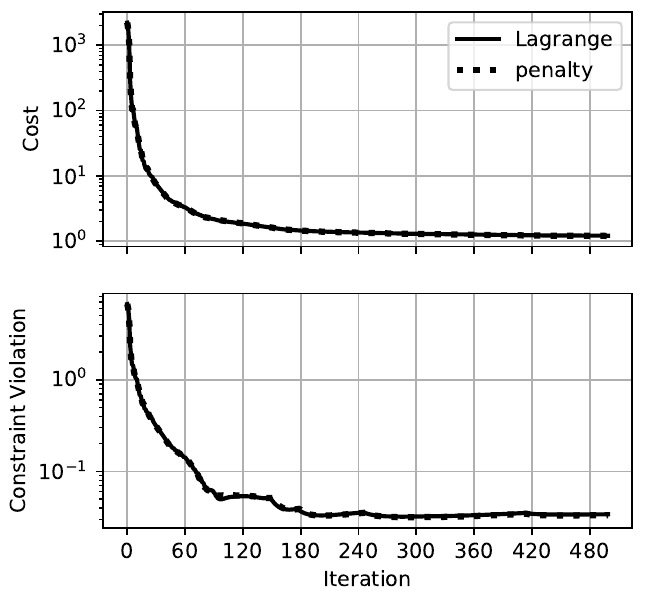}
        \caption{Dual-Arm}
        \label{fig:open_loop_convergence:dual_jaco}
    \end{subfigure}
    \begin{subfigure}{0.47\linewidth}
        \centering
        \includegraphics[width=0.9\linewidth]{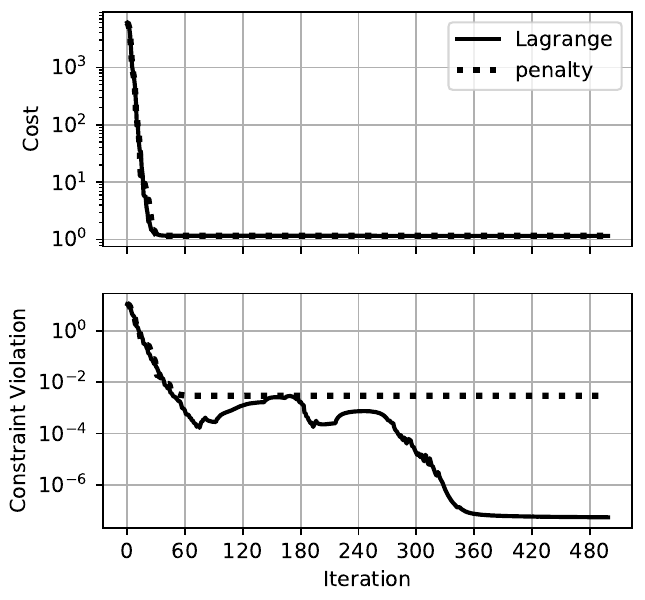}
        \caption{Allegro Hand}
        \label{fig:open_loop_convergence:allegro}
    \end{subfigure}
    \caption{Convergence with quadratic penalty and Lagrange multipliers for
        each of the four examples. }
    \label{fig:open_loop_convergence}
\end{figure*}

For each of the examples described above, we performed open-loop optimization
with both the penalty method and the Lagrange multipliers (LM) method (see
Section~\ref{sec:solver:equality_constraints}). Convergence plots are shown in
Fig.~\ref{fig:open_loop_convergence}. This figure shows the cost and constraint
violations (sum of squared generalized forces on unactuated DoFs) at each 
iteration. Since underactuation constraints are the only dynamics constraints in
IDTO, constraint violations provide a measure of dynamic feasibility.

The LM method sometimes performed much better than the penalty method (spinner
and Allegro hand), and occasionally found higher-quality local minima. This is
shown in the spinner example, where the penalty was not sufficient to drive the
finger to touch the spinner, leading to a high cost and large constraint
violations.

\begin{figure}
    \centering
    \includegraphics[width=\linewidth]{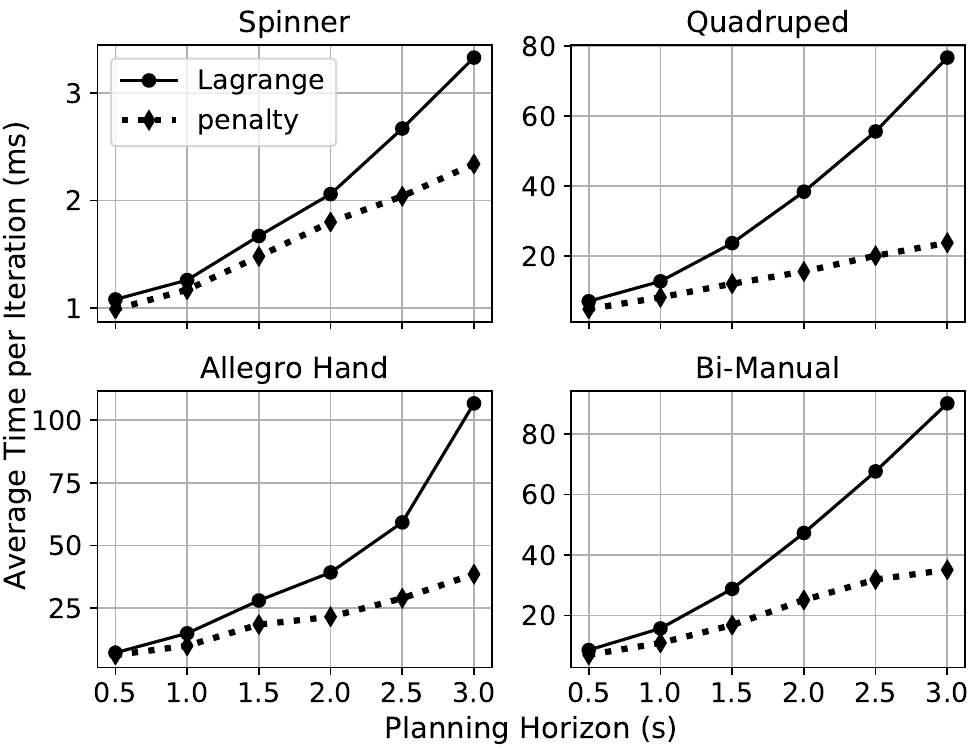}
    \caption{Average wall-clock time per iteration for various planning
        horizons. All examples used a 50~ms time step, with cost and gradient
        computations parallelized across 4 threads. }
    \label{fig:iter_time_vs_horizon}
\end{figure}

For the quadruped and bi-manual manipulation cases, the two methods produced
nearly identical results. In these cases, the penalty method is preferable due
to the cost of solving $\blambda$ from (\ref{eq:lagrange_multiplier}). This
additional cost is illustrated in Fig.~\ref{fig:iter_time_vs_horizon}, which
plots the average wall-clock time per iteration for different planning horizons.
With the penalty method, complexity is linear in the planning horizon and cubic
in the number of DoFs, similar to iLQR/DDP. Scalability is worse with the LM
method: our current implementation solves (\ref{eq:lagrange_multiplier}) using
dense algebra, rendering complexity cubic in the planning horizon. 

We emphasize that the convergence shown in Fig.~\ref{fig:open_loop_convergence}
is not particularly good: for many of the examples, constraint violations are
still large and the cost is still slowly decreasing even after 500 iterations.
However, we found these solutions to be informative in offline CITO
computations, particularly with the quick user interaction cycles enabled by the
speed of these computations. For MPC, we found that high control rates
compensate for poor convergence accuracy.

\begin{figure}
    \centering
    \includegraphics[width=\linewidth]{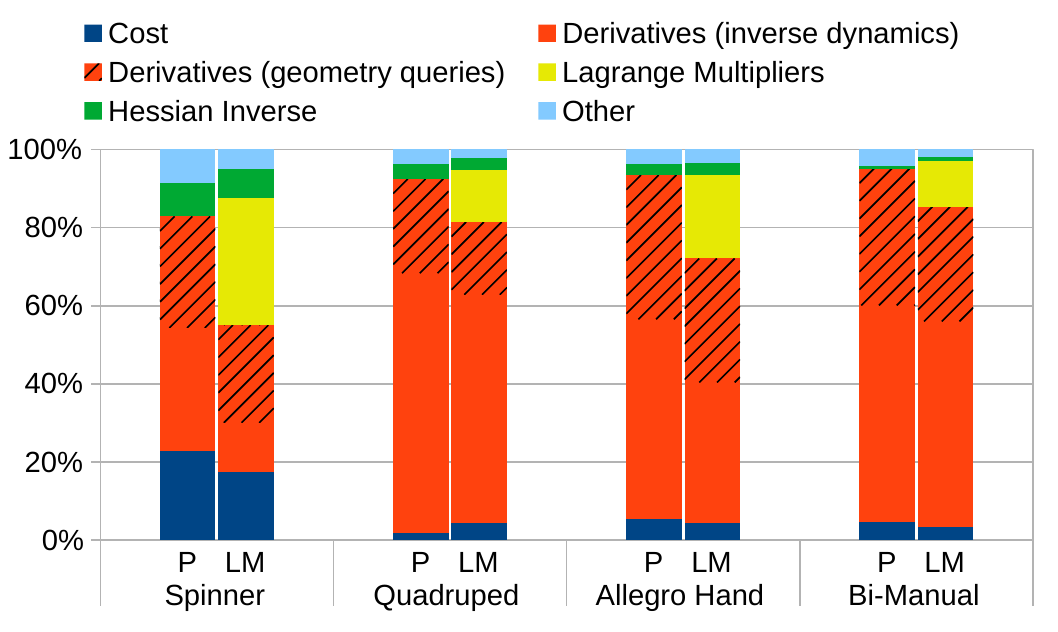}
    \caption{Percent of CPU time dedicated to major computational elements,
        using the penalty method (P) or Lagrange multipliers (LM) to handle equality
        constraints. Derivative and cost calculations (blue, orange, striped orange)
        can be trivially parallelized.}
    \label{fig:profiling}
\end{figure}

Figure~\ref{fig:profiling} shows a breakdown of the computational cost, using a
2~s (40 step) horizon for all of the examples. Inverse dynamics derivatives
(orange) are the most expensive component. Fortunately, derivatives and cost
calculations are easily parallelized.

\hl{Table~\ref{tab:param_sensitivity} illustrates the sensitivity of our solver
to various contact parameterizations. Specifically, it shows the constraint
violation magnitude after 500 iterations, a rough measure of performance, on the
spinner problem for various contact stiffness ($k$) and stiction velocity
($v_s$) values.

For our default stiction velocity of $v_s = 0.05$ m/s (top row), the solver
performs well at stiffnesses up to $2 \cdot 10^8$ N/m. This exceeds the
stiffness of steel, which has a stiffness on the order of $10^7$ N/m
\citep{castro2023theory}. However, as we decrease the stiction velocity $v_s$,
the performance of the solver drops off, particularly for large values of $k$.
Smaller stiction velocities correspond to a closer approximation of coulomb
friction. Like high stiffness, low stiction velocity provides a closer
approximation of rigid contact at the price of a worse-conditioned trajectory
optimization problem. This motivates our choice of contact parameters that are
as soft as possible while maintaining good closed-loop performance, as described
in Section~\ref{sec:formulation:contact:params}.

}

\begin{table}[h]
    \small\sf\centering
    \caption{\hl{Spinner constraint violation after 500 iterations, under various contact stiffnesses $k$ and stiction velocities $v_s$.}}
    \label{tab:param_sensitivity}
    \hl{
    \begin{tabular}{c|cccl}
        \toprule
        \diagbox[height=2em, width=4em]{$v_s$}{$k$} & \msf{2 \cdot 10^2} & \msf{2 \cdot 10^4} & \msf{2 \cdot 10^6} & \msf{2 \cdot 10^8} \\
        \midrule
        \msf{0.05} &  \msf{6.1 \cdot 10^{-8}} & \msf{2.5 \cdot 10^{-8}} & \msf{7.2 \cdot 10^{-8}} & \msf{2.8 \cdot 10^{-7}}\\
        \msf{0.01} &  \msf{2.8 \cdot 10^{-7}} & \msf{2.3 \cdot 10^{-7}} & \msf{3.0 \cdot 10^{-4}} & \msf{2.8 \cdot 10^{-1}}\\
        \msf{0.005} & \msf{4.2 \cdot 10^{-5}} & \msf{2.7 \cdot 10^{-3}} & \msf{1.0 \cdot 10^{-3}} & \msf{3.2 \cdot 10^{-1}}\\
        \msf{0.001} & \msf{1.3 \cdot 10^{-2}} & \msf{7.2 \cdot 10^{-2}} & \msf{1.1 \cdot 10^{-0}} & \msf{5.0 \cdot 10^{-0}}\\
    \end{tabular}
    }
\end{table}

\subsection{Model Predictive Control}\label{sec:mpc_results}

We applied our solver to CI-MPC for all four example scenarios. \hl{We did not
run the solver to convergence, but rather performed a single iteration at each 
MPC step, as in a real-time iteration scheme \citep{diehl2005real}.}

Between iterations, we tracked the latest solution with a
higher-frequency feed-forward PD controller,
\begin{equation}\label{eq:pd_plus}
    \u = \u_{ff} + \K_P(\q_{d} - \hat{\q}) + \K_D(\v_{d} - \hat{\v}),
\end{equation}
where feedforward torques $\u_{ff}$, desired positions $\q_{d}$, and desired
velocities $\v_{d}$ were obtained from a cubic spline interpolation of IDTO
solutions, with the latest solution as the knot points. $\hat{\q}$ and
$\hat{\v}$ are state estimates, and $\K_P$ and $\K_D$ are gain matrices.

\subsubsection{Spinner}

The goal is to move the spinner 2 radians past its starting position. For MPC,
the starting position was constantly updated to match the current position, such
that the spinner kept spinning. Damping in the spinner joint means that the
robot must constantly interact with the spinner to accomplish this task.

With the LM method and parallelization across 4 threads on a laptop (Intel
i7-6820HQ, 32 GB RAM), MPC ran in real time at about 200~Hz. A ``finger
gaiting'' cycle emerged from the optimization, as shown in the accompanying
video. While the optimizer's contact model allows some force at a distance,
simulations with Drake's contact model do not \citep{castro2020transition}.

The spinner is inspired by an example from \cite{posa2013direct}, which reports
around 30 seconds of offline computation to generate similar behavior.

\subsubsection{Mini Cheetah Quadruped}

As for the spinner, we updated the quadruped's goal at each iteration to move
the robot forward at about 0.4~m/s. Again, while the planner's compliant contact
model allows force at a distance and includes very large friction
regularization, the simulator used Drake's contact model with tight
regularization of friction, physics-based compliance, and no action at a
distance.

With the penalty method and parallelization across 4 threads on a laptop, MPC
ran in real-time at about 60~Hz. Over flat portions of the terrain, a trot-like
gait emerged, with opposite pairs of legs working together. The robot deviated
from this gait to cross two small hills. Screenshots from the generated
trajectory are shown in Fig.~\ref{fig:mini_cheetah}, and the full simulation is
shown in the supplemental video. 

\begin{figure*}
    \centering
    \includegraphics[width=\linewidth]{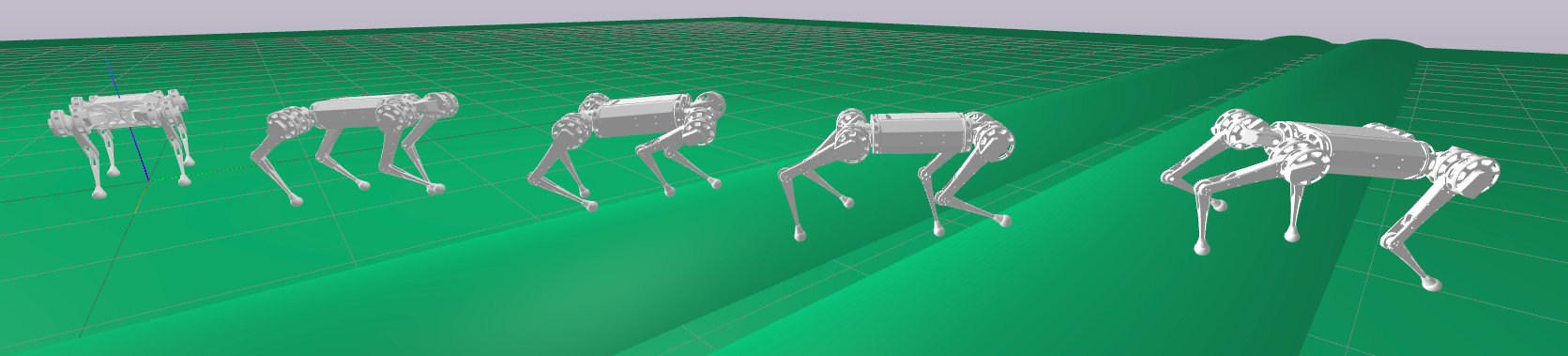}
    \caption{A simulated Mini Cheetah quadruped traverses two small hills. The
        simulation runs in real-time while IDTO acts as an MPC controller at
        about 60~Hz. We do not specify a preferred contact sequence, only a
        quadratic cost defining the desired direction of travel. Our CI-MIPC
        planner uses a relaxed model of compliant contact, while the simulator
        uses Drake's rigid contact model.}
    \label{fig:mini_cheetah}
\end{figure*}

\begin{figure}
    \centering
    \includegraphics[width=0.9\linewidth]{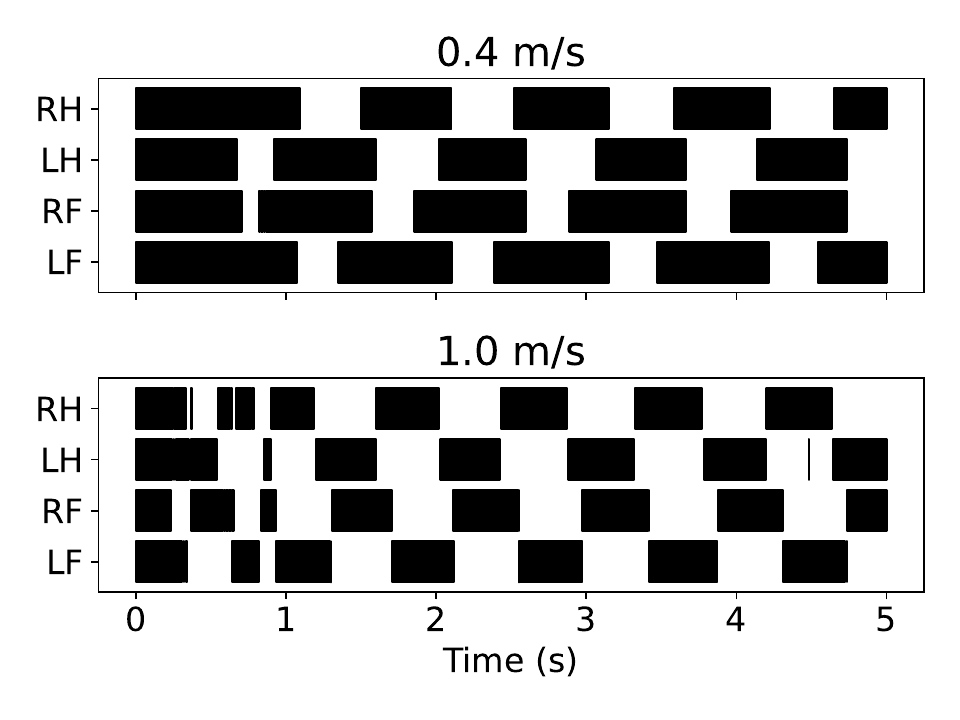}
    \caption{\hl{Contact pattern for Mini Cheetah moving over flat ground at
    different target velocities. Black patches indicate a contact phase for the
    given foot. A trotting gait emerges at various speeds, with IDTO
    automatically adjusting the gait frequency and amount of overlap between
    phases. }}
    \label{fig:gait_diagram}
\end{figure}

\hl{A trotting gait emerged at various speeds over flat ground, as illustrated
in Fig.~\ref{fig:gait_diagram}. In this figure, we varied only the target
velocity, keeping the same contact parameterization and cost function. The gait
frequency and time between touch-down and lift-off events were automatically
adjusted by IDTO.}

\subsubsection{Allegro Dexterous Hand}

The robot was to rotate the ball 180 degrees in its hand. Unlike IDTO's point
contact, the simulation used a hydroelastic model of surface patches
\citep{elandt2019pressure,masterjohn2021discrete} to simulate the rich
interactions between the hand and the ball.

We found that the LM method was essential to obtaining good performance. With
parallelization across 4 threads on a laptop, MPC ran in real-time at 10~Hz. We
found that the robot could perform different rotations merely by changing the
desired orientation: no further parameter tuning was needed.

\begin{figure*}
    \centering
    \includegraphics[width=0.16\linewidth]{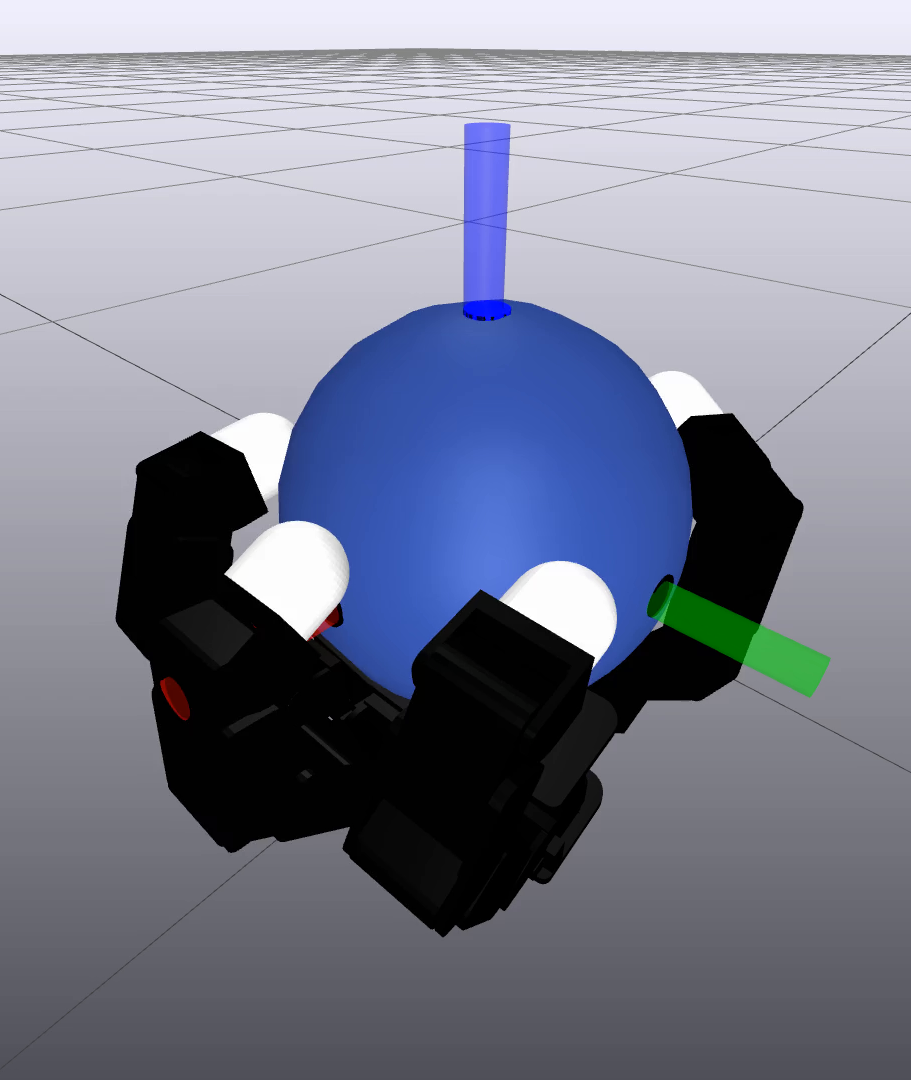}
    \includegraphics[width=0.16\linewidth]{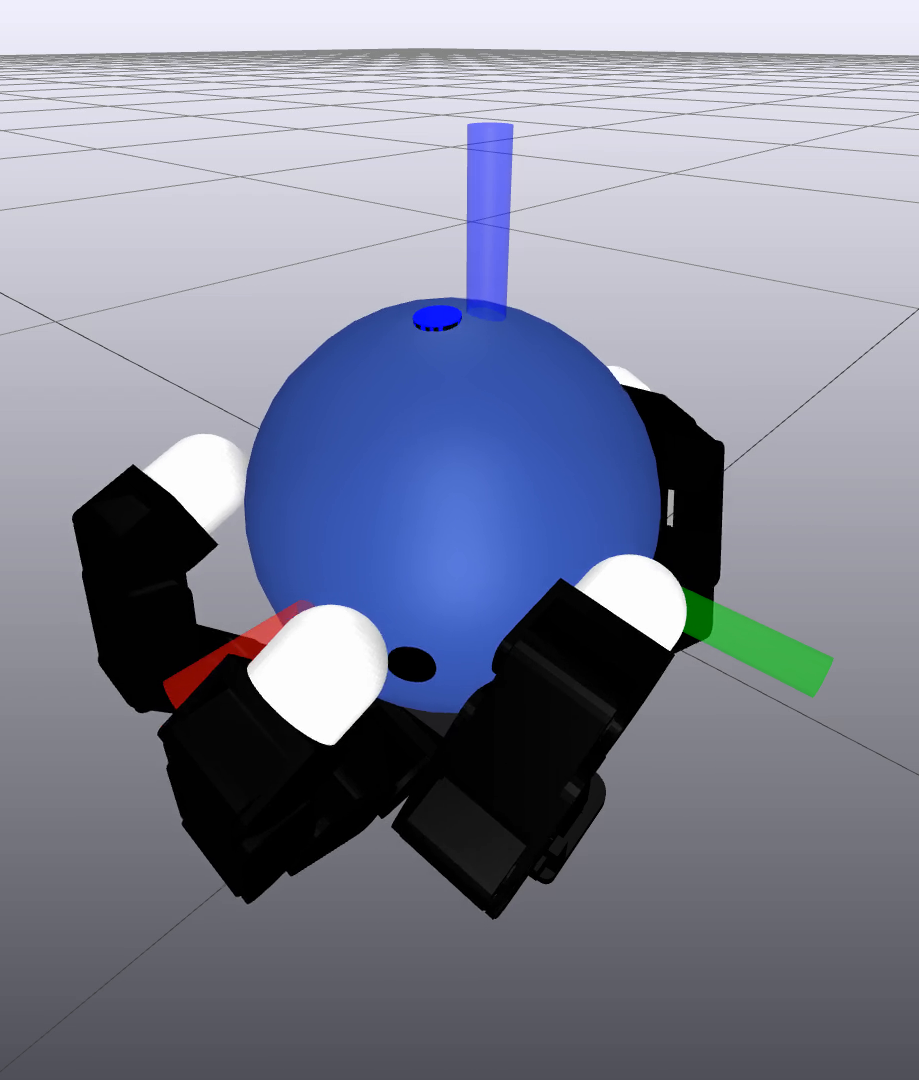}
    \includegraphics[width=0.16\linewidth]{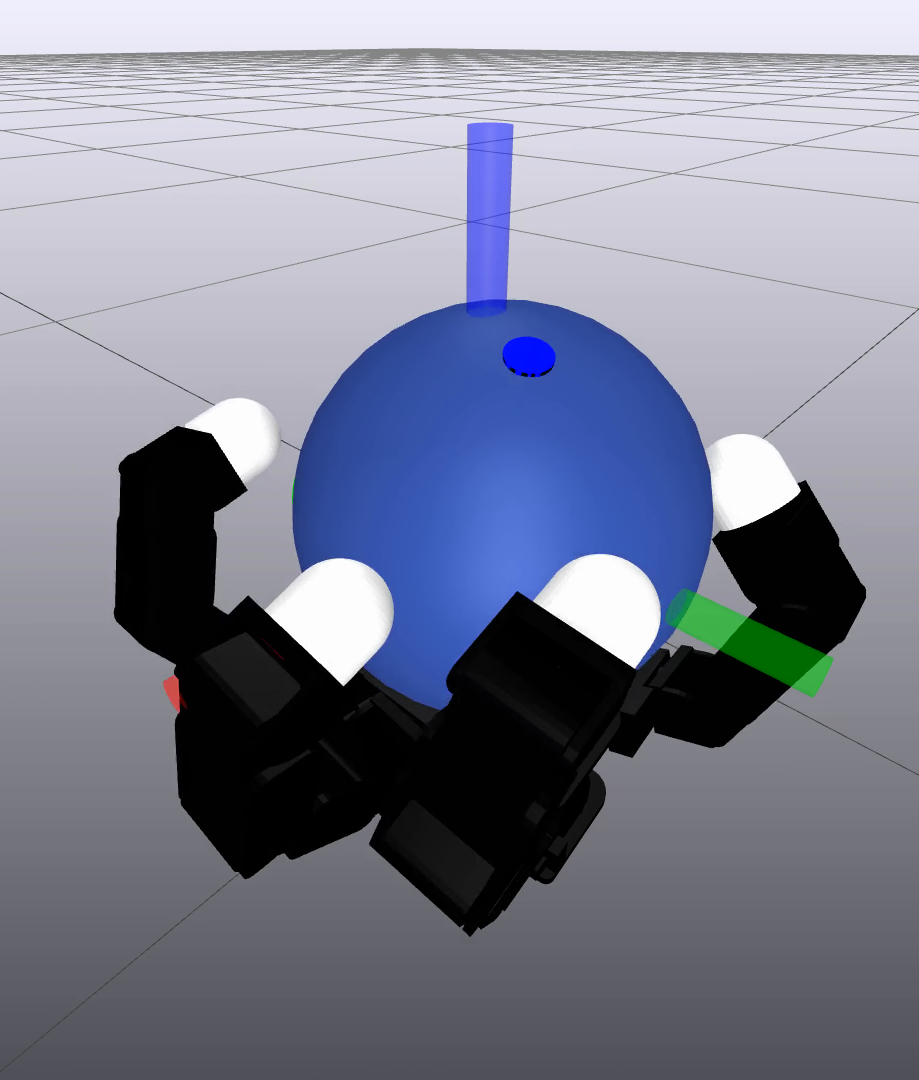}
    \includegraphics[width=0.16\linewidth]{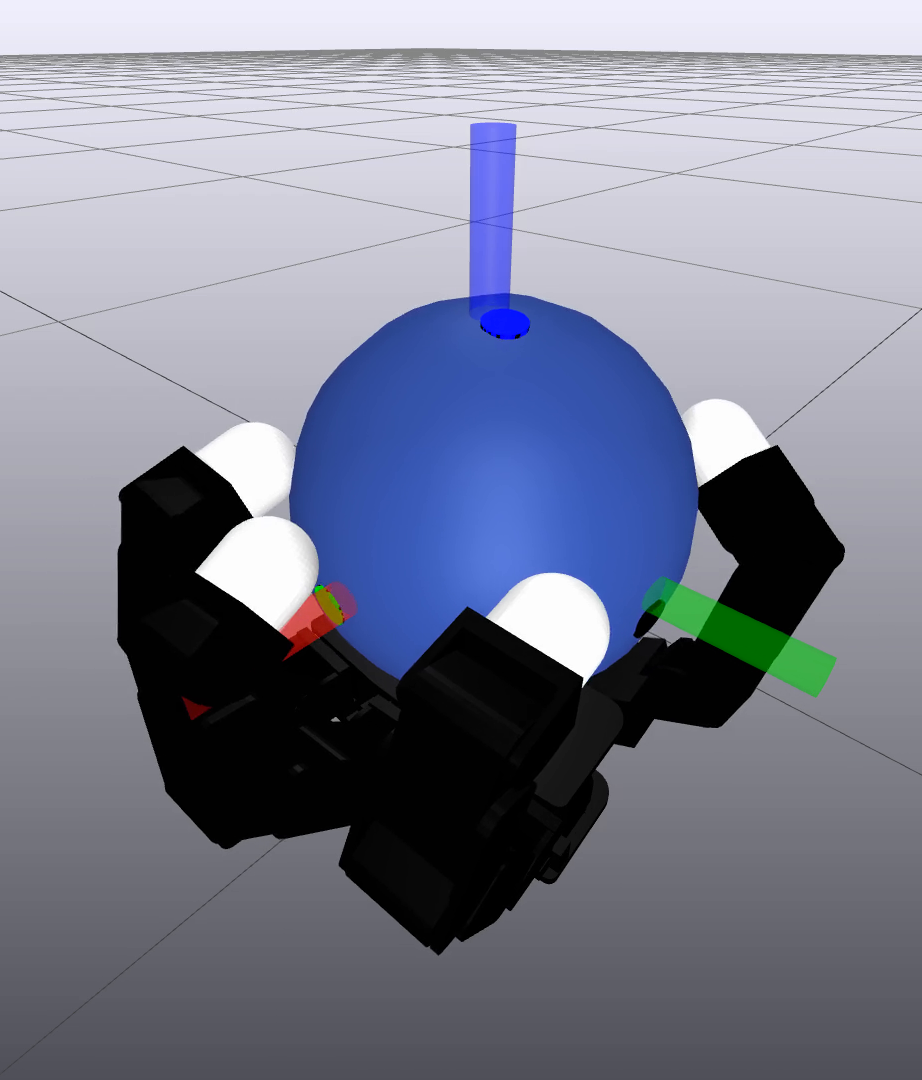}
    \includegraphics[width=0.16\linewidth]{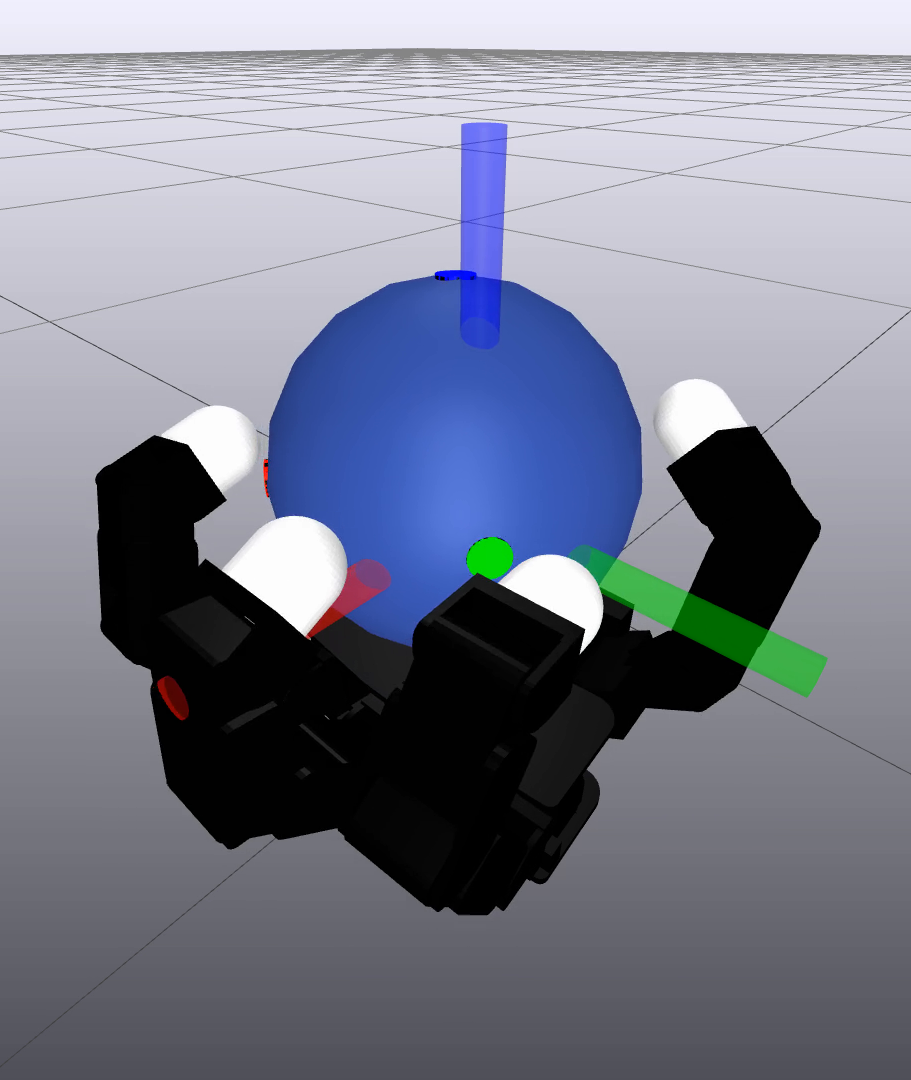}
    \includegraphics[width=0.16\linewidth]{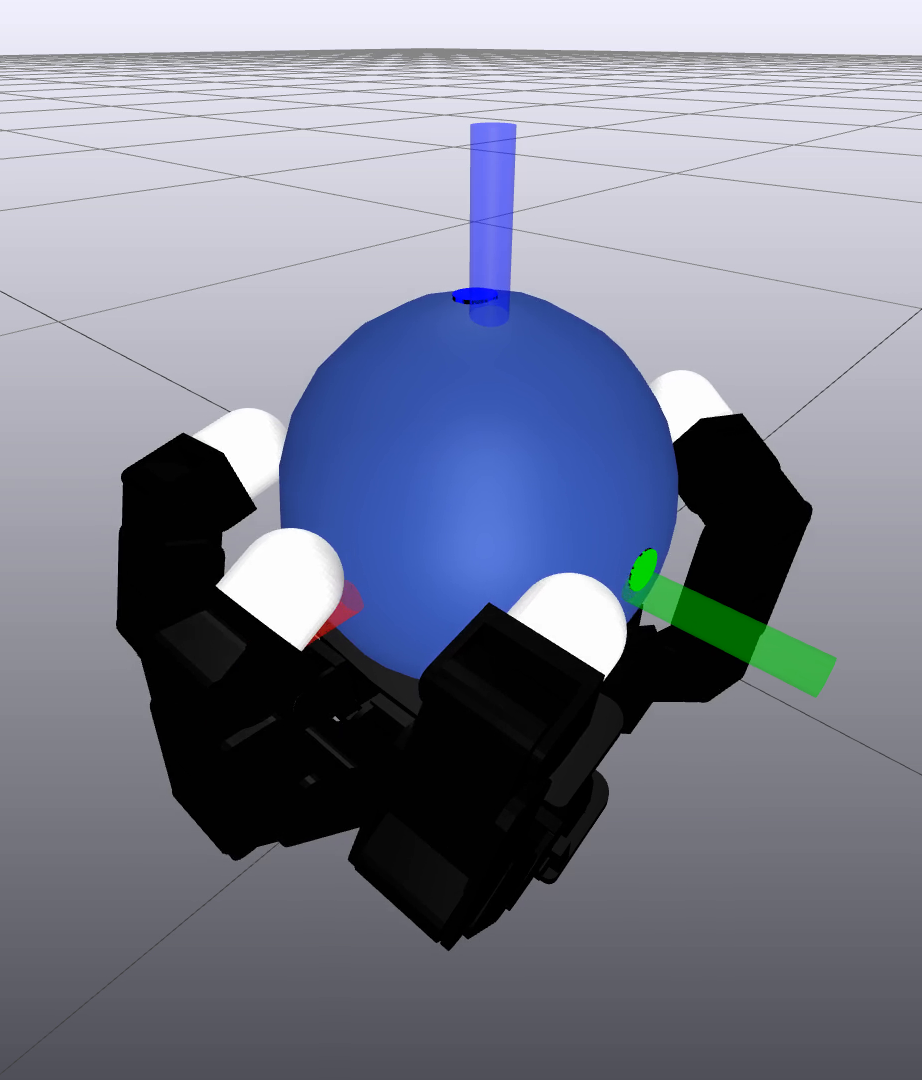}
    \caption{A simulated Allegro dexterous hand rotates a sphere 180 degrees.
        IDTO determines when and where to make and break contact at about 10~Hz.
        The planner uses the simple point contact model described in
        Section~\ref{sec:formulation:contact}, while the simulator uses a more
        realistic hydroelastic model based on contact patches
        \citep{elandt2019pressure,masterjohn2021discrete}. }
    \label{fig:allegro}
\end{figure*}

The robot completed the 180-degree rotation shown in Fig.~\ref{fig:allegro} in
about 15 seconds, without any offline computation. The resulting trajectory is
comparable to that obtained by \citep{pang2022global}, which uses a quasi-dynamic
model and requires around a minute of offline computing to perform a similar
180-degree in-hand rotation.

\subsubsection{Bi-Manual Manipulator}

Finally, we validated our approach on hardware with two Jaco arms. We first
designed several behaviors---push a box on the ground, pick up the box, and
balance the box on its edge---in simulation. We used hydroelastic contact for
the simulation and ran MPC at around 15~Hz on a laptop.

For the hardware experiments, we used a system with a 24-core processor (AMD
Ryzen Threadripper 3960x, 64 GB RAM). We parallelized derivative computations
across 20 threads (one per time step). Together with the fact that the computer
did not need to simultaneously run a simulation, this allowed us to perform MPC
between 100-200~Hz. The exact rate varied through the experiments, depending on
the contact configuration.

\begin{figure*}
    \centering
    \includegraphics[width=0.16\linewidth]{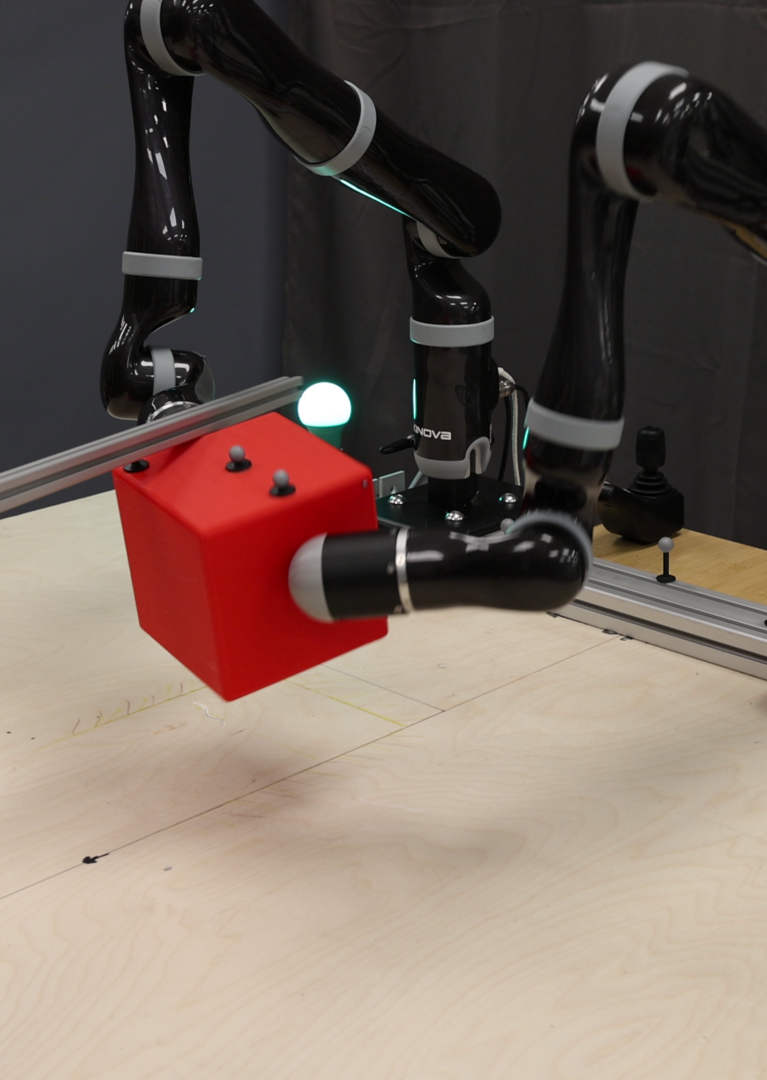}
    \includegraphics[width=0.16\linewidth]{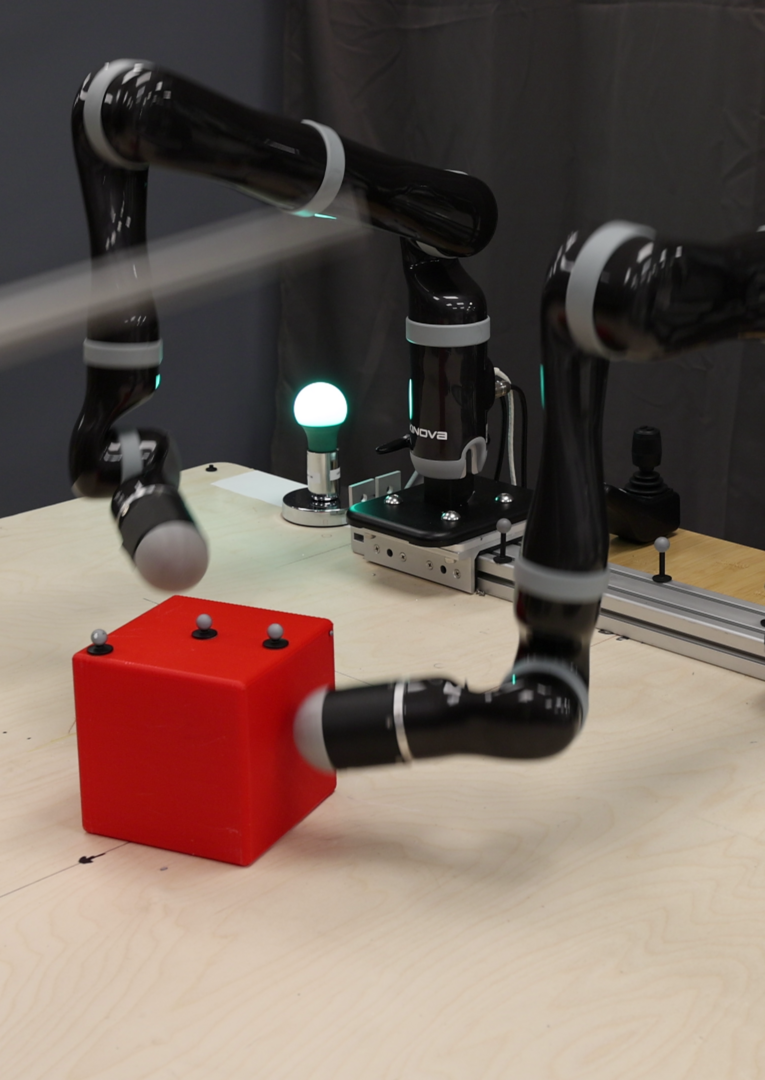}
    \includegraphics[width=0.16\linewidth]{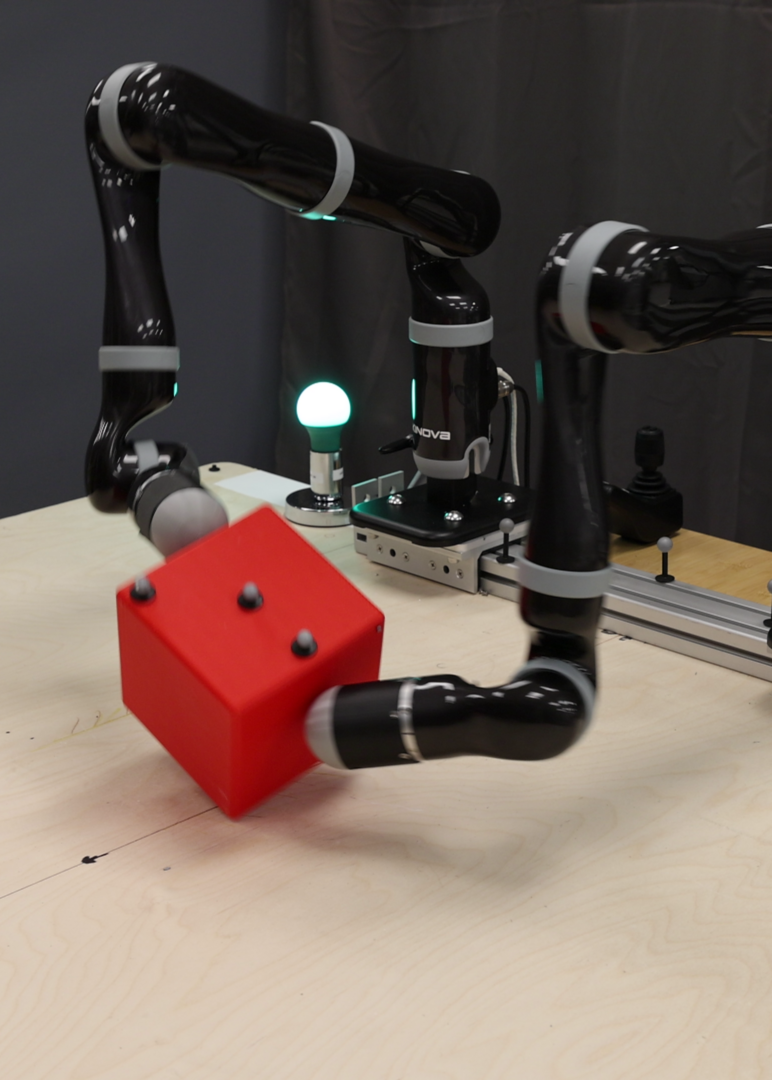}
    \includegraphics[width=0.16\linewidth]{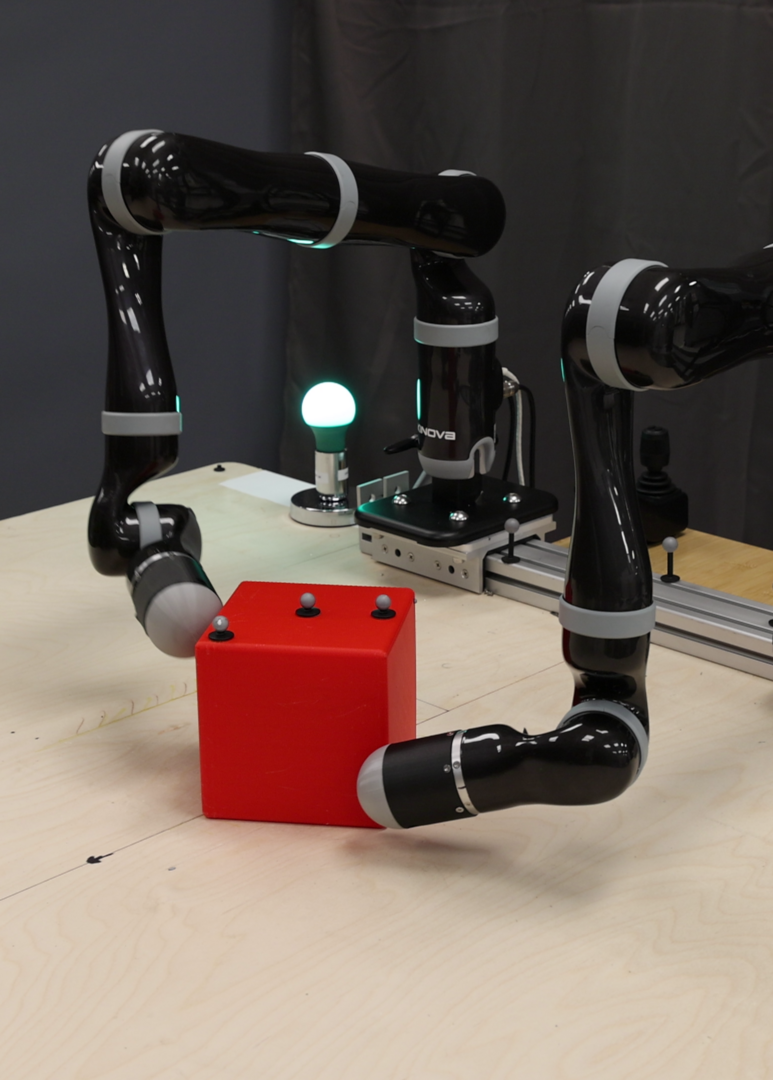}
    \includegraphics[width=0.16\linewidth]{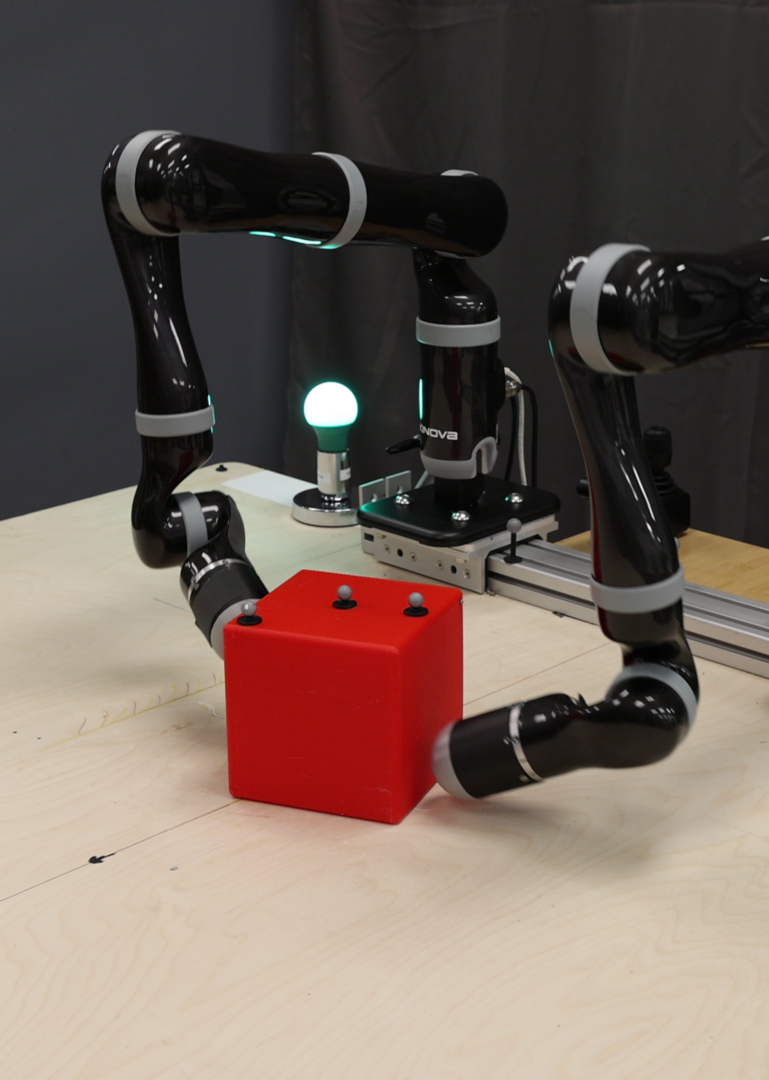}
    \includegraphics[width=0.16\linewidth]{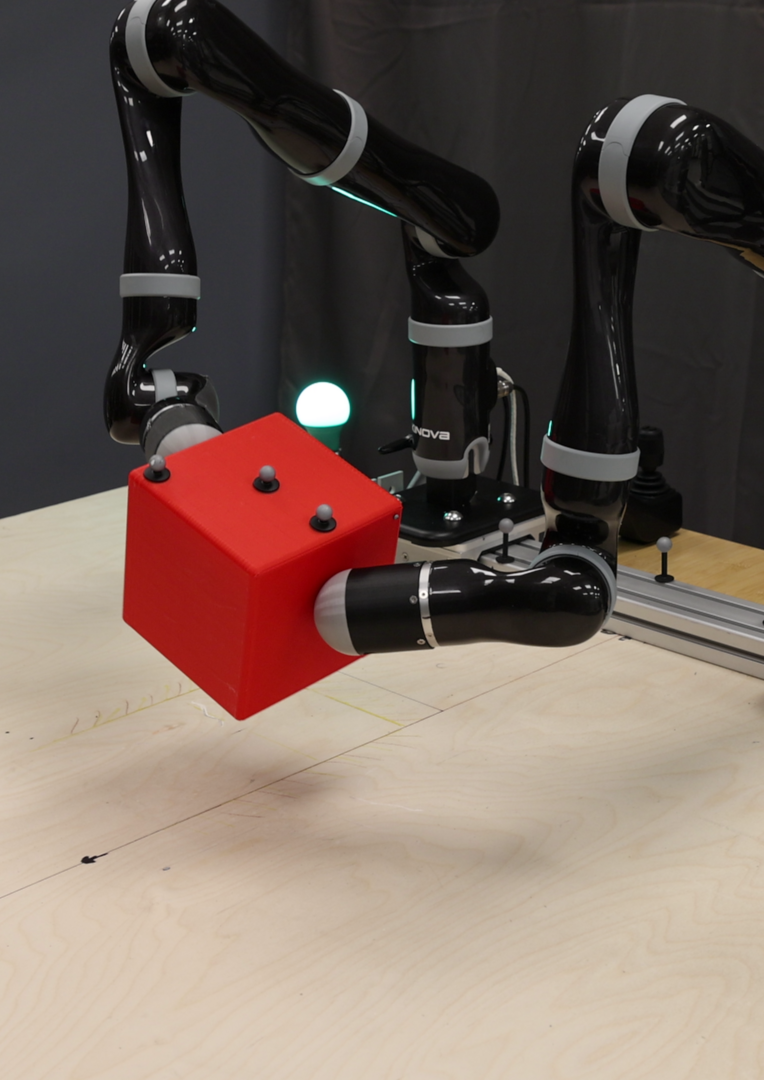}
    \caption{Two 7-DoF Kinova Jaco arms tasked with lifting a box recover from
        an external disturbance (push with a stick from above). After dropping
        the box, the robot attempts to re-grasp but fails. The robot
        then selects a new contact configuration and succeeds on the second
        attempt.}
    \label{fig:hardware_lift}
\end{figure*}

With a few exceptions, we used the same contact parameters in simulation and on
hardware. The exceptions primarily had to do with friction since we did not
have accurate friction measurements available. We also used a larger friction
coefficient (1.0 rather than 0.2) for the planner in the lifting task, as this
encouraged the robot to squeeze the box more gently, and we wanted to push the
box out of the robot's grasp, as shown in Fig.~\ref{fig:hardware_lift}.

The hardware setup differed slightly from that of the simulation. We operated
the robot in position-control mode using Jaco's proprietary controller with
default gains. For the box, an \emph{Optitrack} motion capture system measured
pose, and we assumed zero velocity.

Footage of the hardware experiments can be found in the accompanying video.
Screenshots of the picking-up task are shown in Fig.~\ref{fig:hardware_lift}.
These experiments highlight the usefulness of CI-MPC: our solver was able to
recover from significant external disturbances, making and breaking contact,
changing between sticking and sliding modes and adjusting contact
configurations on the fly.

Our experiments also illustrate some limitations of IDTO, particularly with
respect to local minima. This is especially evident in the pushing example,
where the robot is tasked with pushing the box on the table to a desired pose.
The robot responds quickly and effectively to small disturbances but is not
able to recover from a large disturbance that takes the box far from the arms.
With the box far away, there are no gradients that indicate to the solver that
the arms should reach around the box to push it back: the system is stuck in a
local minimum where the arms do not move.

The severity of such local minima can be reduced by increasing the smoothing
parameter $\sigma$, but there is a tradeoff: with a very large $\sigma$ the
planner expects a considerable amount of force at a distance and may fail to actually make contact. A more problem-specific solution would be to add a cost
terms that encourage the end-effector to move to the side of the manipuland
opposite the target pose \citep{aydinoglu2023consensus}.

\hl{
\subsection{Constraints and Performance}

Our solver often takes many iterations to satisfy constraints to tight
tolerances (Fig.~\ref{fig:open_loop_convergence}), a likely consequence of our
Gauss-Newton Hessian approximation \citep{nocedal1999numerical}. How do
constraint violations impact CI-MPC performance? Is it important that a CI-MPC
solver satisfy constraints to numerical precision?

In this section, we show that tight constraint satisfaction is not critical to
closed-loop CI-MPC performance. We focus in particular on the spinner as a
simple illustrative example.

Figure~\ref{fig:constraints_and_mpc} plots constraint violation over time during
CI-MPC, where a single IDTO iteration is performed at each CI-MPC step. We
compare two cases. In the first, a fixed target configuration $\bar{\q}$ is
used, and constraint violations are eventually driven to zero (dashed lines). In
the second, we constantly shift the target configuration so that the spinner is
always rotating. In this case, constraint violations do not go to zero. Spikes
in constraint violation correspond to contact events, as the finger repeatedly
flicks the spinner.

\begin{figure}
    \centering
    \includegraphics[width=0.8\linewidth]{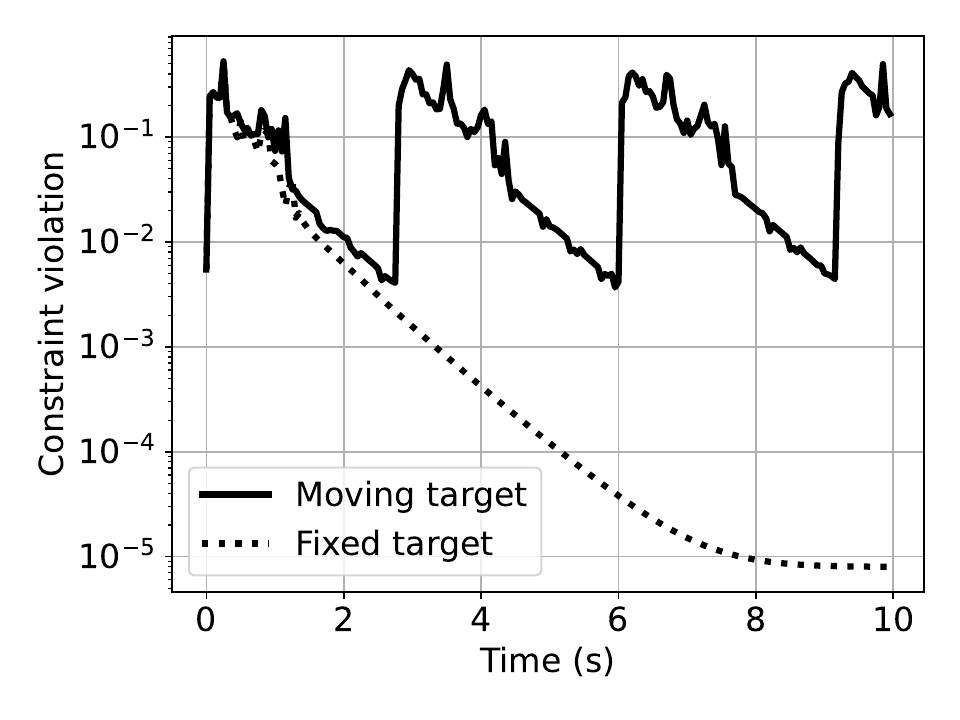}
    \caption{\hl{Constraint violations during closed-loop MPC with the spinner.
    When the target configuration $\bar{\q}$ is fixed, we see convergence over
    time (dotted line). If we continually update the target configuration so the
    spinner is constantly rotating (solid line), constraint violations don't
    grow unbounded, but don't converge either. Both examples correspond to good
    qualitative performance.}}
    \label{fig:constraints_and_mpc}
\end{figure}

Many practical CI-MPC applications are more similar to the second case. Even if
the objective does not shift over time, a changing environment leads to an
optimization landscape that is constantly shifting. This is consistent with the
analogy of ``surfing'' rather than ``mountain climbing'' for CI-MPC
\citep{howell2022predictive}: tight constraint satisfaction and optimality is
less important than fast iteration times and sufficient improvement at each
iteration.

Figure~\ref{fig:constraints_open_loop} plots the spinner angle predicted by IDTO
with an actual simulated rollout at different constraint violation levels.
Unsurprisingly, lower constraint violations correspond to a closer match between
the predicted and actual trajectories. However, even when constraint violations
are quite large (top subplot), the predicted trajectory roughly captures the key
aspects of the system's behavior: in this case, the fact that the spinner
rotates once the finger pushes it. 

Similarly, IDTO's predictions are not perfect even at convergence (bottom
subfigure). This is due to a mismatch between the contact models and
integration schemes of the simulator and the planner. Persistent prediction
error even at convergence further emphasizes the fact that tight constraint
satisfaction is not critical, a result consistent with recent results in the
literature \citep{khazoom2024tailoring}.

\begin{figure}
    \centering
    \includegraphics[width=\linewidth]{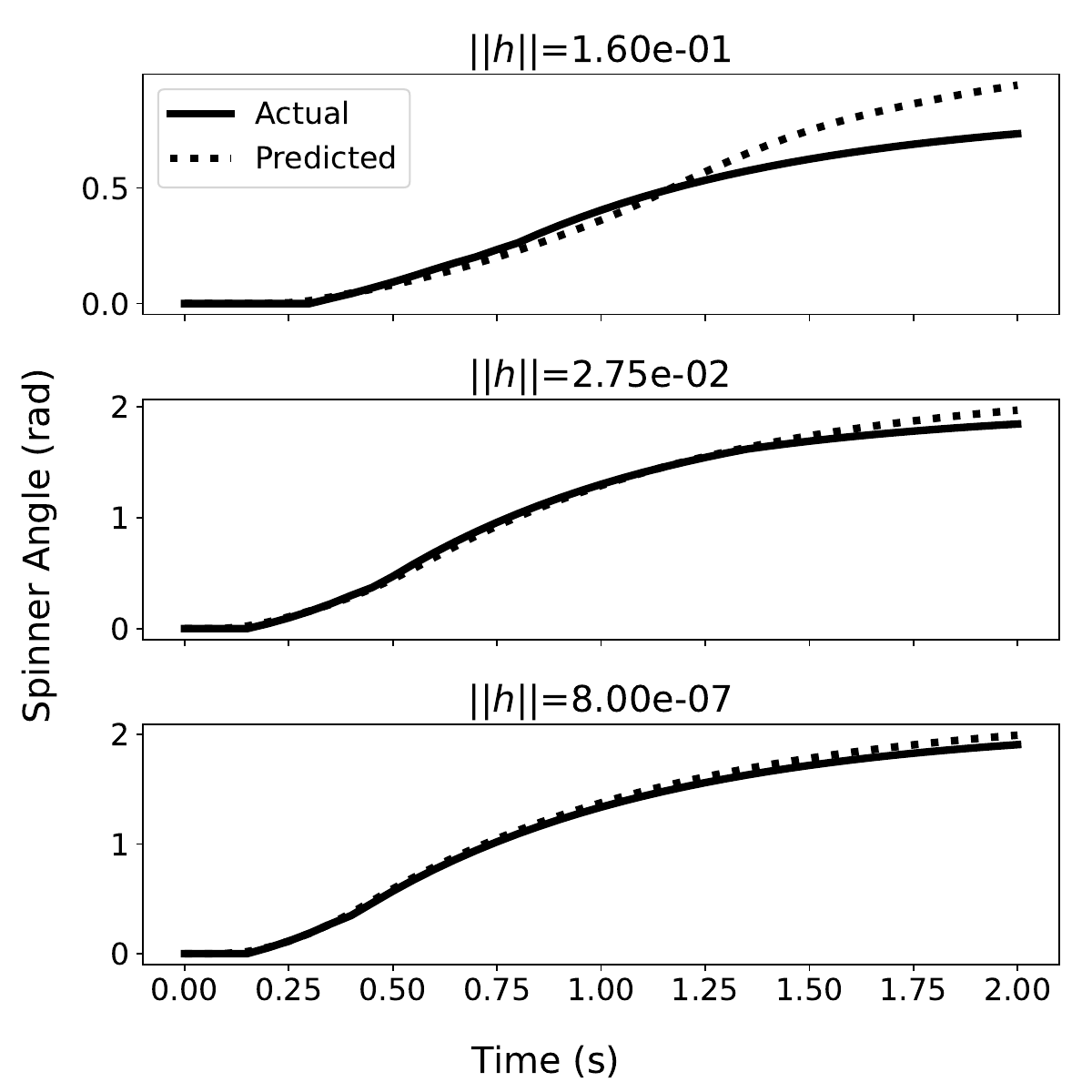}
    \caption{\hl{Comparison between the spinner angle predicted by MPC (dotted lines)
    and the actual trajectory when MPC's controls are applied open-loop (solid
    lines), across various constraint violation levels. IDTO's predictions capture
    the key features of the system's behavior, even when constraint violations
    are large (top).
    }}
    \label{fig:constraints_open_loop}
\end{figure}

Because tight constraint satisfaction is less critical than fast iteration
times, we generally prefer the simple quadratic penalty method
(Section~\ref{sec:solver:penalty}) to the Lagrange multipliers method
(Section~\ref{sec:solver:lagrange}) for handling constraints. While
there are some cases where Lagrange multipliers can reduce the severity of local
minima, the cost of slower iteration times is rarely worth it. As a result, we
recommend that users start with the penalty method, and only switch to the
Lagrange multipliers method if constraint violations pose a challenge or if it
reduces local minima on a particular problem.

}
\section{Comparison with Existing Work}\label{sec:comparison}

Table~\ref{tab:realtime_mpc_comparison} compares IDTO to results reported in the
literature. For brevity, we focus on recent results that run CI-MPC in real
time. Because CI-MPC performance often depends heavily on implementation details
and parameter tuning, we choose to compare with results reported in the
literature rather than attempting to reimplement the methods ourselves. We
believe this fascilitates the fairest possible comparison, as authors are
incentivised to report their best results.

We encourage the reader to read the cited references for further details. Notice
that the oldest work is from 2018, with most from 2022 or newer, so major
performance differences cannot be attributed to differences in processing power.
Nonetheless, hardware and software quality do significantly impact performance,
and most methods in Table~\ref{tab:realtime_mpc_comparison} would perform better
with further optimization. This includes our own approach: while Drake is
currently undergoing significant performance improvements, IDTO would benefit
from faster multibody algebra. For example, MuJoCo can simulate a humanoid with
4000 timesteps per second on a single CPU \citep{howell2022predictive}. By
contrast, Drake runs a comparably sized model at around 800 timesteps per
second.

\begin{table*}
    \small\sf\centering
    \caption{Comparison with real-time CI-MPC in the literature. Data marked with ? was not reported.}
    \label{tab:realtime_mpc_comparison}
    {\footnotesize
    \begin{tabular}{cccccccccc}
        \toprule
        Method                                      & System       & DoFs & Horizon (s) & $\delta t$ (s) & MPC Iters. & Freq.  &
        \begin{tabular}{@{}c@{}}Simplified \\ Dynamics\end{tabular}                   &
        \begin{tabular}{@{}c@{}}Pref. Contact \\ Sequence\end{tabular}                   &
        Hardware                                                                                                                                                          \\
        \midrule
        \cite{neunert2018whole}         & Quadruped    & 18   & 0.5         & 0.004          & 1          & 190 Hz & \textbf{No} & Yes         & \textbf{Yes} \\
        \cite{kong2022hybrid}              & Quadruped    & 18   & 0.5         & 0.01           & 1          & ?      & \textbf{No} & Yes         & \textbf{Yes} \\
        \cite{howell2022predictive}      & Quadruped    & 18   & 0.25        & 0.01           &1 & 100 Hz & \textbf{No} & Yes         & No           \\
        \cite{howell2022predictive}      & Shadow Hand  & 26   & 0.25        & 0.01           &1 & 100 Hz & \textbf{No} & \textbf{No} & No           \\
        \cite{cleac2023fast}        & Push Bot     & 2    & 1.6         & 0.04           & ?          & 70 Hz  & \textbf{No} & \textbf{No} & No           \\
        \cite{cleac2023fast}        & Quadruped    & 18   & 0.15        & 0.05           & ?          & 100 Hz & Yes         & Yes         & \textbf{Yes} \\
        \cite{aydinoglu2023consensus} & Finger Pivot & 5    & 0.1         & 0.01           & 5          & 45 Hz  & \textbf{No} & \textbf{No} & No           \\
        \cite{aydinoglu2023consensus} & Ball Roll    & 9    & 0.5         & 0.1            & 2          & 80 Hz  & Yes         & \textbf{No} & \textbf{Yes} \\
        \cite{kim2023contact}               & Quadruped    & 18   & 0.5         & 0.025          & 4          & 40 Hz  & \textbf{No} & Yes         & \textbf{Yes} \\
        IDTO (ours)                                 & Spinner      & 3    & 2.0         & 0.05           & 1          & 200 Hz & \textbf{No} & \textbf{No} & No           \\
        IDTO (ours)                                 & Hopper       & 5    & 2.0         & 0.05           & 1          & 100 Hz & \textbf{No} & \textbf{No} & No           \\
        IDTO (ours)                                 & Quadruped    & 18   & 1.0         & 0.05           & 1          & 60 Hz  & \textbf{No} & \textbf{No} & No           \\
        IDTO (ours)                                 & Bi-Manual    & 20   & 1.0         & 0.05           & 1          & 100 Hz & \textbf{No} & \textbf{No} & \textbf{Yes} \\
        IDTO (ours)                                 & Allegro Hand & 22   & 2.0         & 0.05           & 1          & 10 Hz  & \textbf{No} & \textbf{No} & No           \\
        \bottomrule
    \end{tabular}}
\end{table*}

Table~\ref{tab:realtime_mpc_comparison} shows that IDTO allows longer planning
horizons than many existing methods, a fact that is enabled by relatively
large time steps. We do not simplify the system dynamics, though we do simplify
collision geometries (see Remark~\ref{remark:collision_geometries}). We are
particularly proud of the fact that IDTO does not require a preferred contact
sequence for the quadruped, though specifying one is possible in the
IDTO framework and might improve performance in practice.

\hl{
\subsection{Benefits of Inverse Dynamics}

In this section, we compare the performance of IDTO and a state-of-the-art
forward-dynamcics-based CI-MPC method: the contact-implicit iLQR implementation
of MuJoCo MPC \cite{howell2022predictive}. While an accurate head-to-head
comparison is difficult due to the complexity of CI-MPC methods and the
importance of implementation details, we believe the fact that MuJoCo MPC also
offers a performant and well-maintained C++ implementation makes for as fair a
comparison as possible. 

We focus on the spinner example, for which both methods obtain good closed-loop
performance. While this example is relatively simple, it captures two key
benefits of IDTO over a forward dynamics formulation: faster iteration times and
reduced susceptibility to local minima.

\subsubsection{Faster Iteration Times}

First, note that the MuJoCo dynamcis backend used by the iLQR baseline is
consistently faster than the Drake dynamics backend used by our IDTO solver.
This is shown in Fig.~\ref{fig:sim_speeds}, where we compare the time to
simulate the spinner open loop for 2.0 seconds with random control actions. We
use a time step of $0.05$ seconds for both methods, and average over 10,000
trials. Mean and standard deviation are shown for both an older laptop (i7 CPU)
and a newer desktop (i9 CPU). This provides a significant baseline advantage to
iLQR, as it is implemented over a faster dynamics backend. 

\begin{figure}
    \centering
    \begin{subfigure}{\linewidth}
        \centering
        \includegraphics[width=\linewidth]{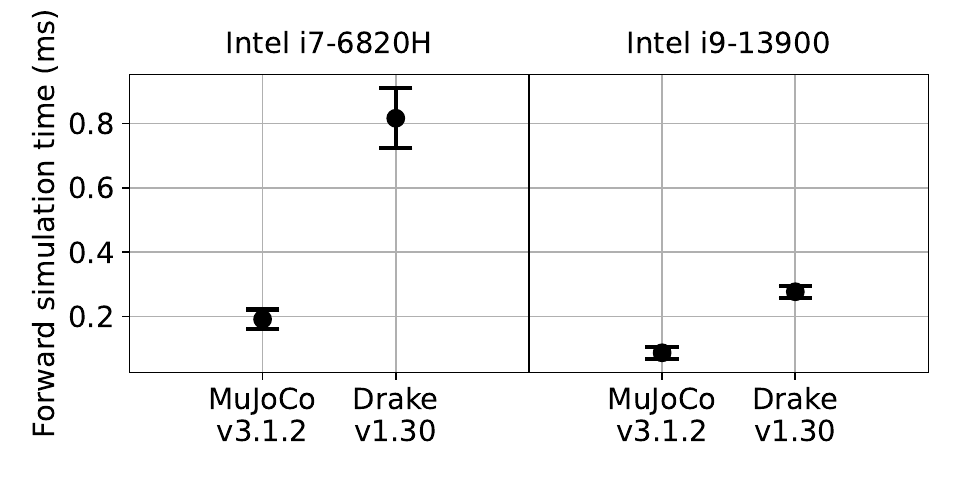}
        \caption{\hl{Time to simulate the spinner for 2.0
        seconds with random control actions. MuJoCo is consistently faster than
        Drake.}}
        \label{fig:sim_speeds}
    \end{subfigure}
    \begin{subfigure}{\linewidth}
        \includegraphics[width=\linewidth]{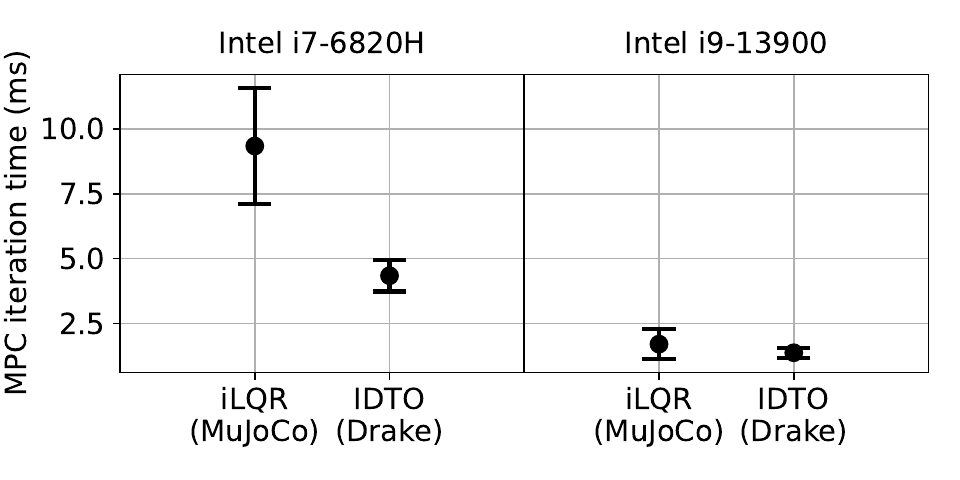}
        \caption{\hl{MPC iteration times for the spinner with a 2.0 second
        planning horizon. IDTO is generally faster than forward-dynamics-based
        iLQR.}}
        \label{fig:mpc_speeds}
    \end{subfigure}
    \caption{\hl{Iteration times for our IDTO implementation are on par with or
    faster than the forward-dynamics-based iLQR of \cite{howell2022predictive}
    (\subref{fig:mpc_speeds}), despite using a slower dynamics backend
    (\subref{fig:sim_speeds}). The trends are consistent across an older laptop
    (i7 CPU) and a newer desktop (i9 CPU).}}
    \label{fig:speed_comparisons}
\end{figure}

Nonetheless, IDTO generally achieves faster iteration times than iLQR, as shown
in Fig.~\ref{fig:mpc_speeds}. We use the same cost function and planning horizon
(2.0 seconds with a 0.05 s time step) for both methods, and average over 10,000
iterations. IDTO is faster because iLQR requires a full forward simulation at
each iteration, resolving contact dynamics and other details to high precision
with an optimization sub-problem at each time step. In contrast, IDTO only
requires simpler inverse dynamics computations, and does not involve an inner
optimization problem.

\subsubsection{Fewer Local Minima}

Another advantage of IDTO is reduced sensitivity to local minima. To illustrate
this, we compare the closed-loop performance of IDTO and iLQR on two variations
of the spinner problem (Fig.~\ref{fig:ilqr_vs_idto_convergence}). In the first
case (top subplot), the target configuration $\bar{\q}$ puts the finger in
contact with the spinner. Both IDTO and iLQR successfully rotate the spinner to
the desired angle in this case. On the second (bottom subplot), the target
configuration $\bar{\q}$ puts the finger a short distance away from the spinner.
In this case, IDTO successfully rotates the spinner, while iLQR is stuck in a
local minimum where the spinner does not move.

\begin{figure}
    \centering
    \includegraphics[width=\linewidth]{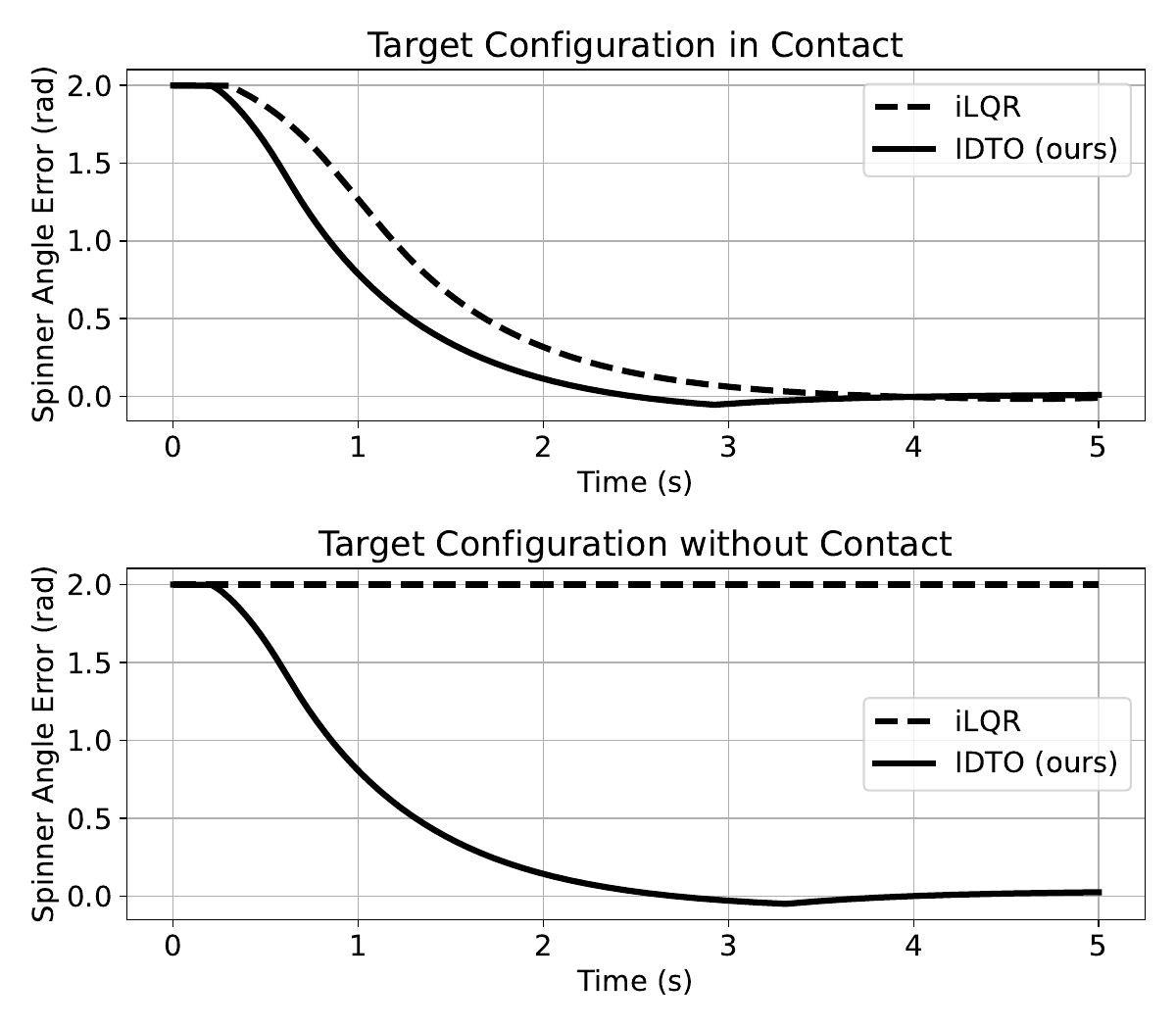}
    \caption{\hl{IDTO reduces the severity of local minima compared to 
    (forward-dynamics-based) iLQR. Both methods are able to drive the spinner to
    the desired angle when the target configuration $\bar{\q}$ involves contact
    between the finger and the side of the spinner (top). However, when the
    target configuration places the finger out of contact, iLQR
    gets stuck in a local minimum where the spinner does not move (bottom).}}
        
    \label{fig:ilqr_vs_idto_convergence}
\end{figure}

While some of this reduction in local minima severity can be attributed to our
contact model's allowance for some force at a distance, there are also more
fundamental reasons that IDTO reduces local minima, even without
force at a distance. As an example of this, consider a simple case of a block
resting on a rigid platform, as shown in Fig.~\ref{fig:block_on_table}.

\begin{figure}
    \centering
    \includegraphics[width=0.48\linewidth]{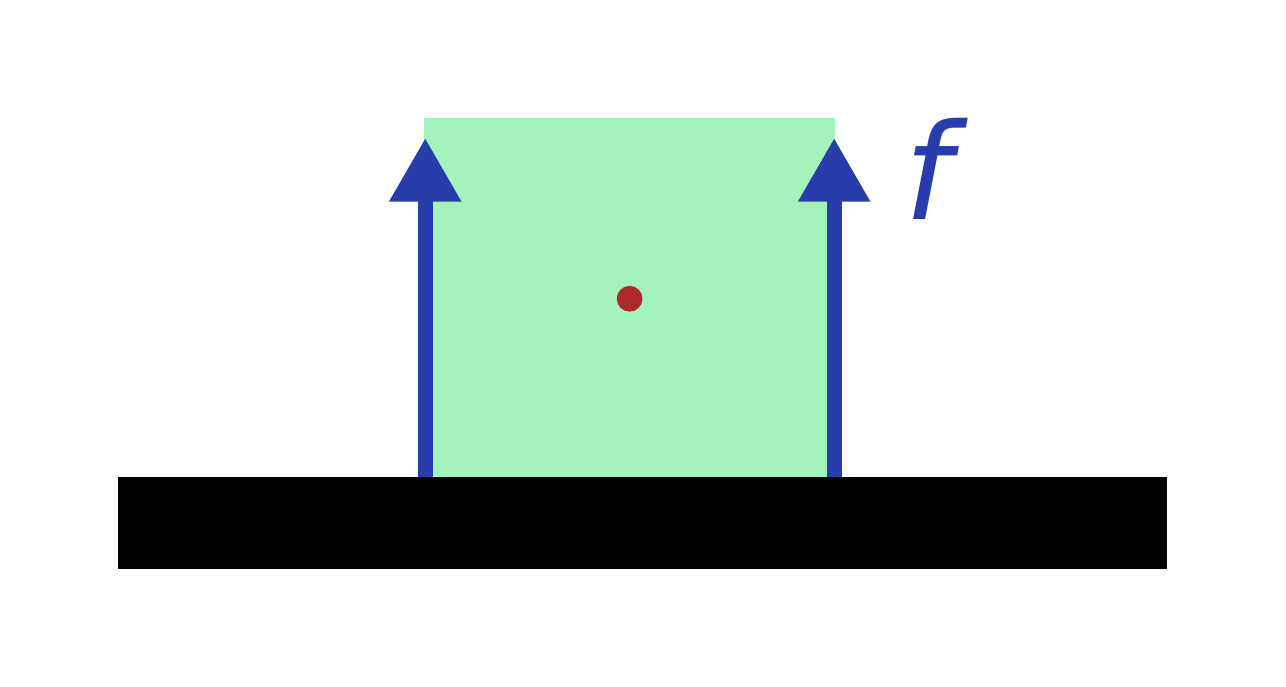}
    \includegraphics[width=0.48\linewidth]{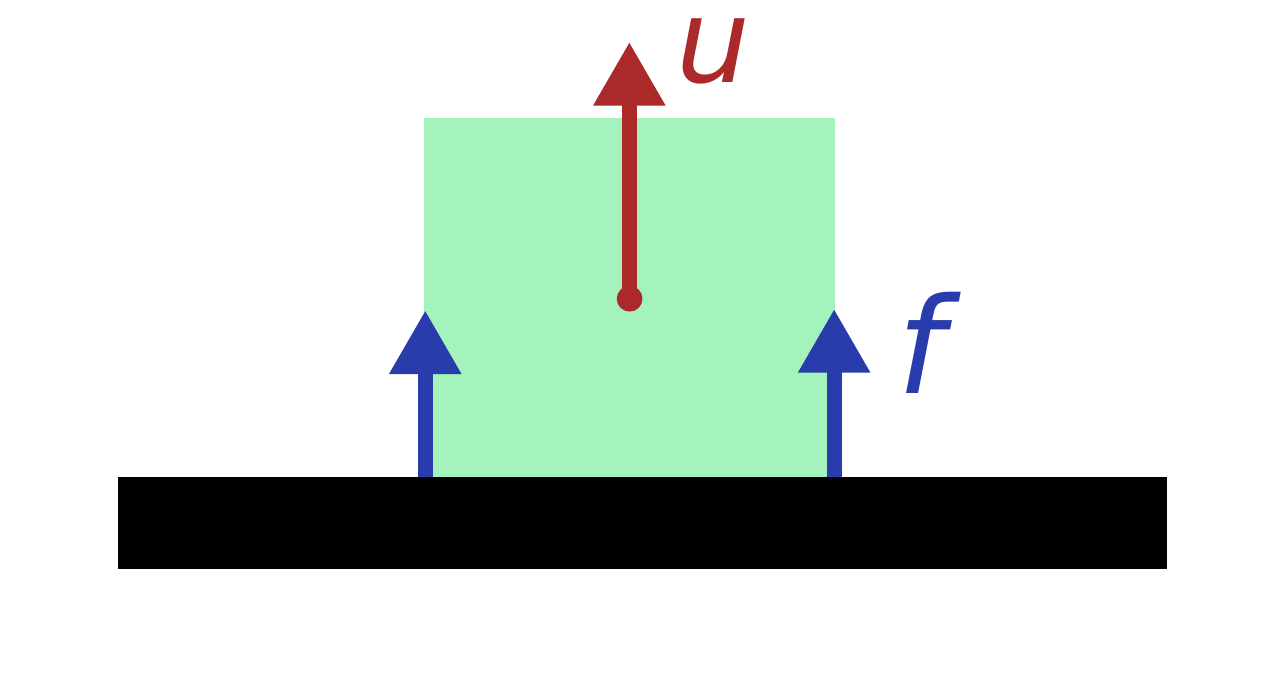}
    \caption{\hl{A block rests on a rigid platform. The task is to apply control
    forces $\u$ to lift the block. For forward-dynamics-based CI-MPC this is a
    local minimum, since small control inputs $\u$ do not change the block's
    position \citep{le2024leveraging}. An inverse dynamics formulation, where
    the decision variables are $\q$, does not contain this local minimum: there
    is a clear gradient showing that the optimal behavior is to lift the box.}}
    \label{fig:block_on_table}
\end{figure}

The block is actuated by a force acting on its center of mass, and the task is
to move upward. With a forward dynamics formulation, the system is stuck in a
local minimum, as applying a small force has no effect on the block's position
\cite{le2024leveraging}. In contrast, under an inverse dynamics formulation,
there are clearly defined gradients with respect to the configuration $\q$,
eliminating the local minimum.
}

\section{Limitations}\label{sec:limitations}

In this section, we highlight the weaknesses of our approach, with an eye
toward future CI-MPC research. 

First, IDTO only guarantees dynamic feasibility at convergence due to the
constraint (\ref{eq:underactuation_constraint}). In practice, non-converged
solutions include forces on unactuated DoFs. For example, the spinner might spin
without anything pushing it. With the quadratic penalty method these
non-physical forces can exist even at convergence. The LM method helps, but does
not eliminate the problem completely. In particular, LM is more costly
than the penalty method and reduces the impact of parallelization
(Fig.~\ref{fig:profiling}). Furthermore, equality constraints are still only
enforced at convergence. LM performance could likely be improved with
sparse algebra in the computation of \eqref{eq:lagrange_multiplier}.

Solver convergence can be slow, as shown in
Fig.~\ref{fig:open_loop_convergence}. This is an inevitable result of the
Gauss-Newton approximation --- problem (\ref{eq:id_traj_opt}) is a large-residual
problem, for which Gauss-Newton methods converge linearly
\citep{nocedal1999numerical}. Accelerating convergence with second-order
techniques is a potentially fruitful area for future research.

Our solver does not currently support arbitrary constraints, such as input
torque or joint angle limits. While approximating such constraints with a
penalty method would be straightforward, this may not be suitable for athletic
behaviors at the limit of a robot's capabilities.

Computationally, the biggest bottleneck is computing derivatives. This is shown
in orange and striped orange in Fig.~\ref{fig:profiling}. Recent advances in
analytical derivatives of rigid-body dynamics algorithms
\citep{singh2022efficient,carpentier2019pinocchio} could potentially alleviate
this bottleneck, but such results would first need to be extended to include
differentiation through contact\hl{, and in particular to account for the
dependence of the jacobian on contact location}.

Practically, our solver's reliance on spherical collision geometries, as
discussed in Remark~\ref{remark:collision_geometries}, presents a major
limitation. While inscribed spheres enabled basic box manipulation on hardware,
this approximation introduces modeling errors that could be problematic for more
complex tasks. A better long-term solution could be a model like hydroelastics
\citep{elandt2019pressure,masterjohn2021discrete} that avoids discontinuous
artifacts for objects with sharp edges, \hl{or a method that leverages the KKT
conditions of the optimization problem for the signed distance function to
obtain a smooth approximation \citep{dietz2024high}}.

As for most CITO solvers, cost weights and contact parameters determine IDTO's
performance in practice. While we found that IDTO was not particularly sensitive
to changes in the cost weights, as evidenced by the multiples of 10 in
Table~\ref{tab:cost_weights}, IDTO is sensitive to contact modeling parameters,
particularly the stiction velocity \hl{(Table~\ref{tab:param_sensitivity})}.
Nonetheless, we found that a fast solver made contact parameter tuning easier.
This could be further improved with a graphical interface
\citep{howell2022predictive}.

Finally, IDTO is a fundamentally local optimization method and as such is
vulnerable to local minima. A more systematic ``virtual force'' framework
\citep{onol2020tuning,todorov2019optico} could help reduce the severity of these
local minima. \hl{Integration with a higher-level global planner is another
promising avenue toward better performance. The fact that local optimization
with IDTO is sufficient for tasks as complex as generating quadruped gaits and
in-hand object rotation with a dexterous hand presents an opportunity to reduce
the burden on a higher-level global planner, which might merely give some
suggestion of promising contact areas rather than specifying a complete contact
sequence.}

\section{Conclusion}\label{sec:conclusion}

IDTO is a simple but surprisingly effective tool for planning and control
through contact. Even with a relaxed contact model and simple quadratic costs,
IDTO enables real-time CI-MPC in simulation (with rigid contact) and on
hardware, and is competitive with the state of the art. We believe that the
simplicity of IDTO is responsible for much of its performance, and hope that it
can provide a foundation for further improvements in the future.

Areas for future research include improving contact discovery with virtual
forces, developing faster (analytical) derivatives for inverse dynamics with
contact, considering convex contact models \citep{castro2022unconstrained}, and
extracting a local feedback policy in the style of iLQR/DDP.
\hl{While we chose an implicit integration scheme out of a desire for numerical
stability, higher-accuracy semi-implicit integration schemes are compatible with
the IDTO framework, and could improve performance.} Further improvements could
stem from proper handling of $SE(3)$ structure in the dynamics and second-order
methods for faster convergence.

Finally, to mitigate the local nature of IDTO,
our solver could be combined with a higher-level global planner based on
sampling \citep{pang2022global}, graph search \citep{natarajan2023torque},
learning \citep{chi2023diffusion}, or combinatorial motion planning
\citep{marcucci2021shortest}.

\begin{acks}
    Thanks to Sam Creasey for helping with the Jaco hardware, Jarod
    Wilson for 3D printing parts, Andrew Patrikalakis for setting up an external
    repository, and the Tactile/Punyo and Dynamics/Simulation teams at Toyota
    Research Institute for their resources and support.
\end{acks}
\appendix

\section*{Appendix}

\subsection*{Computing the Hessian Approximation}\label{apx:gauss_newton_hessian}

As outlined in Section \ref{sec:solver:gauss_newton}, we compute a Gauss-Newton
approximation of the Hessian. This approximation neglects second-order time
derivatives of the inverse dynamics when propagating derivatives through
\eqref{eq:explicit_cost_gradient}.

Separating the weight matrix $\Q = \text{diag}([\Q_{\q}~\Q_{\v}])$
into position and velocity components and noting the symmetry of $\H$, we
compute nonzero blocks as follows:

\footnotesize
\begin{equation*}
    \H_{k,k-2} = 
        \left[\frac{\partial \btau_{k-1}}{\partial \q_{k-2}}\right]^T\R\frac{\partial \btau_{k-1}}{\partial \q_{k}},
\end{equation*}
\begin{multline*}
    \H_{k,k-1} = 
        \left[\frac{\partial \v_t}{\partial \q_{k-1}}\right]^T\Q_{\v}\frac{\partial \v_t}{\partial \q_{k}} \\
        + \left[\frac{\partial \btau_{k-1}}{\partial \q_{k-1}}\right]^T\R\frac{\partial \btau_{k-1}}{\partial \q_{k}}
        + \left[\frac{\partial \btau_{k}}{\partial \q_{k-1}}\right]^T\R\frac{\partial \btau_{k}}{\partial \q_{k}},
\end{multline*}
\begin{multline*}
    \H_{k,k} = \Q_{\q}  + 
        \left[\frac{\partial \v_t}{\partial \q_t}\right]^T\Q_{\v}\frac{\partial \v_t}{\partial \q_t}
        + \left[\frac{\partial \v_{k+1}}{\partial \q_t}\right]^T\Q_{\v}\frac{\partial \v_{k+1}}{\partial \q_t} + \\
        \left[\frac{\partial \btau_{k-1}}{\partial \q_t}\right]^T\R\frac{\partial \btau_{k-1}}{\partial \q_{k}}
        + \left[\frac{\partial \btau_{k}}{\partial \q_t}\right]^T\R\frac{\partial \btau_{k}}{\partial \q_{k}}
        + \left[\frac{\partial \btau_{k+1}}{\partial \q_t}\right]^T\R\frac{\partial \btau_{k+1}}{\partial \q_{k}}.
\end{multline*}
\normalsize

Constructing this Hessian approximation requires only velocity and inverse
dynamics derivatives --- $\frac{\partial \v_k}{\partial \q_k}$, $\frac{\partial
\v_{k+1}}{\partial \q_k}$, $\frac{\partial \btau_{k-1}}{\partial \q_k}$,
$\frac{\partial \btau_{k}}{\partial \q_k}$, and $\frac{\partial
\btau_{k+1}}{\partial \q_k}$ --- as discussed in
Section~\ref{sec:solver:gauss_newton}. Our solver calculates these derivatives
once per iteration and caches them for computational efficiency.

\bibliographystyle{SageH}
\bibliography{references}

\end{document}